%% file: task_oriented_trans.tex
\RequirePackage[2024-11-01]{latexrelease}
\documentclass[journal,english]{IEEEtran}

\usepackage{comment}

\usepackage{epsfig}
\usepackage{iftex}
\RequirePDFTeX
\usepackage[T1]{fontenc}
\usepackage{times}
\usepackage{float}
\usepackage{algorithm}
\usepackage{algpseudocode}

\algrenewcommand{\algorithmicrequire}{\textbf{Input:}}
\algrenewcommand{\algorithmicensure}{\textbf{Output:}}
\usepackage{stfloats}
\usepackage{balance}
\usepackage{afterpage}
\usepackage{placeins}
\makeatletter
\newcommand{\nextresultpage}{\newpage\if@firstcolumn\else\newpage\fi}
\makeatother

\usepackage{amsmath}
\usepackage{amstext,cite}
\usepackage{amssymb,bm}
\usepackage{latexsym}
\usepackage{xcolor}
\usepackage{enumitem}
\usepackage{graphicx}
\usepackage{amsthm}
\usepackage[caption=false,font=footnotesize]{subfig}
\usepackage{booktabs}
\usepackage{multicol}
\usepackage{lipsum}

\usepackage[hidelinks]{hyperref}
\usepackage{aecompl}
\usepackage{bbm}
\usepackage{pifont}

\usepackage{empheq}
\usepackage{mathrsfs}
\allowdisplaybreaks

\input{macros}

\def\BibTeX{{\rm B\kern-.05em{\sc i\kern-.025em b}\kern-.08em
		T\kern-.1667em\lower.7ex\hbox{E}\kern-.125emX}}

\theoremstyle{remark}
\renewcommand{\topfraction}{0.95}

\renewcommand{\textfraction}{0.05}

\makeatletter
\g@addto@macro\normalsize{%
  \abovedisplayskip 3.5pt plus 2pt minus 1pt
  \belowdisplayskip 3.5pt plus 2pt minus 1pt
  \abovedisplayshortskip 0pt plus 2pt
  \belowdisplayshortskip 3.5pt plus 2pt minus 1pt
}
\makeatother
\begin{document}
	
	\title{Learning Nonlinear Waveform Fusion Across Scales for Integrated Sensing and Communication}
	\author{
		Jiayu~Liu, 
		Kai~Wan,~\IEEEmembership{Member,~IEEE,} 
		Shuo~Shao,~\IEEEmembership{Member,~IEEE,}
		and~Giuseppe Caire,~\IEEEmembership{Fellow,~IEEE}}
	\date{}
	\maketitle
\begin{abstract}
Fixed source superposition and frequency partitioning constrain how
communication content interacts with a probing reference in integrated
sensing and communication (ISAC). We learn nonlinear waveform fusion across scales for image transmission and localization of multiple targets in range and angle. Reception uses the information available at each end. Adaptive temporal
feature weighting and spectral updates combine image features and a
deterministic linearly frequency-modulated reference in a shared complex
state, generating a waveform with fixed total energy over the array frame.
At the user equipment (UE), the reference and channel state information
condition a neural demixer for image reconstruction. At the base station,
the realized transmit pulse defines a coherent dictionary for sequential
localization with learned local correction, physical refinement, and
residual updates. We compare the proposed waveforms with superposition in the time domain and learned spectral linear fusion on
512 independent test images, using a shared fixed UE receiver
and a baseline sensing receiver using coherent peak search.
Averaged across eleven task weights, the proposed waveforms
reduce normalized localization error by 29.0\% and 27.0\%,
respectively, while increasing mean peak signal-to-noise
ratio by 1.40 and 0.82~dB. With the proposed waveforms
fixed, learned local correction reduces mean normalized localization
error by 12.2\% relative to the same sequential receiver without
learned correction.
\end{abstract}


\section{Introduction}
\label{sec:intro}

Integrated sensing and communication (ISAC) uses a shared radio platform
for information delivery and environmental
sensing~\cite{liu2020jointreview,liu2022integrated}. A shared waveform must deliver recoverable information to a communication
terminal and illuminate targets for parameter estimation~\cite{zhang2021overview,zhou2022integrated}. For image transmission and target localization, we optimize
the shared waveform directly in terms of image reconstruction error and errors in range and angle.

Designs centered on communication reuse the structure of communication
signals for sensing. Orthogonal frequency-division multiplexing (OFDM)
supports radar processing through known symbols
and subcarriers~\cite{tigrek2012ofdm}; adaptive designs adjust their
allocation to communication and sensing
requirements~\cite{liu2017adaptive}. Designs centered on sensing instead embed
information into signals designed for radar. Sidelobe control and waveform diversity preserve selected probing
properties while carrying data~\cite{hassanien2015dual}, and chirp
modulation provides another mechanism for embedding information
~\cite{bekar2020joint}. Tunable filter design controls sidelobe levels,
range sidelobe modulation, and signal-to-noise ratio (SNR) loss for waveforms integrating radar and communication~\cite{jiang2020tunable}. In these approaches,
prescribed signal structures guide information embedding and radar
processing.

Joint optimization adjusts waveform and spatial degrees of freedom.
Multiple-input multiple-output (MIMO) radar and MIMO communication designs serving multiple users optimize transmission and beamforming for both
functions~\cite{liu2018mu,liu2020joint}. Criteria based on beampatterns and estimation bounds connect sensing quality to communication
constraints~\cite{liu2018toward,liu2021cramer}. Analyses of the Cram\'er--Rao bound (CRB) and communication rate
characterize the attainable tradeoffs for specified Gaussian and MIMO
models~\cite{xiong2023fundamental,hua2023mimo}, while symbol-level precoding
also uses instantaneous data symbols in waveform
design~\cite{liu2021dual}. These formulations provide flexible waveform parameterizations, but their objectives at the signal level need not coincide with reconstruction and
localization error.

Learning for physical layer design optimizes transmitter and receiver
mappings from data~\cite{o2017introduction}, including pulse and
constellation shaping~\cite{ait2021end}. Deep joint source--channel coding
maps images to complex channel inputs under a reconstruction
objective~\cite{bourtsoulatze2019deep}. ISAC models for multiple targets learn
steering dictionaries or parameters describing hardware impairments through
differentiable sparse recovery and communication/localization losses to
compensate for array mismatch~\cite{mateos2025model}. Learning also supports
symbol-level precoding~\cite{zheng2024end}. LISAC maps bits nonlinearly to coded OFDM waveforms using weighted communication and sensing proxy losses~\cite{bian2025lisac}; its neural decoder recovers bits, while sensing
uses transmitted symbols in a maximum-likelihood metric for delay and Doppler.
These studies learn different transceiver components;
our focus is the interaction between features carrying image information
and a probing reference during waveform synthesis.

Superposition in the time domain (TD), frequency-division multiplexing
(FDM)~\cite{tagliaferri2023integrated,mao2022waveform} and learned spectral
linear fusion retain a fixed linear rule for source interaction before
normalization. TD adds separately formed sources, FDM assigns complementary
frequency supports, and spectral linear fusion learns bandwise source
transforms. At fixed physical context, changing the image changes the
communication features but leaves the operators incorporating the two
sources unchanged, even when the encoder and fusion parameters are learned.
We ask whether allowing the interaction rule to depend on the current
source features can improve image reconstruction and localization
simultaneously.

To address this question, we learn a mapping that depends on its inputs and couples features carrying image information and a deterministic
linearly frequency-modulated (LFM) reference in a shared
complex state. Temporal convolutions with different fast-time
dilation factors extract multiscale features, while statistics of magnitude and relative phase determine their interaction
weights. Bandwise spectral updates derived from the initial
source features enter the same state before subsequent
temporal processing.

Proposed and the linear controls are optimized
separately under a common energy budget and fixed receive
processing. A temporal nonlinear control assesses the
incremental contribution of the spectral path.

We consider image transmission and localization of static targets
in a monostatic downlink ISAC system. The transmitter uses
communication channel state information (CSI) and a prescribed
sensing covariance, without access to the realized target coordinates.
Nonlinear fusion need not leave the LFM reference as a separately subtractable
component. At the user equipment (UE), the reference and CSI therefore
condition image recovery. At the base station
(BS), the realized transmit pulse defines the coherent dictionary and the
local peak geometry. A local network corrects initialization on a coarse grid
using this geometry and observed correlations. Multiple targets motivate
sequential search, amplitude refitting and residual updates, with physical
refinement of the active coordinates before the next search. 

\subsection{Main Contributions}

We propose an ISAC framework for image transmission and localization of multiple targets in range and angle under a total energy constraint on the array frame. It combines waveform fusion across scales with reception using available waveform information.
The main contributions are as follows.

\begin{enumerate}
\item \textbf{Nonlinear waveform fusion across scales.}
We learn interactions that adapt to current image and LFM features in a shared complex state under a total energy constraint on the array frame. Multiscale temporal convolutions use feature weights based on magnitude and relative phase; bandwise increments from the
initial source features enter the joint state before subsequent
\mbox{temporal processing}.

\item \textbf{Receiver design using available waveform information.}
At the UE, the LFM reference and CSI condition neural demixing for
image reconstruction. At the BS, the realized pulse defines the coherent
dictionary and point-spread features. A shared local network uses these
features and residual correlations to correct candidate initialization
within sequential localization of multiple targets with amplitude refitting,
physical refinement, and residual updates.

\item \textbf{Learning for both tasks and controlled evaluation.}
We optimize normalized losses for image reconstruction and localization in range and angle under fixed receive processing, then adapt the local sensing network
with transmitters fixed. Independent tests averaged over eleven task
weights show higher peak signal-to-noise ratio (PSNR) and lower
normalized localization error than each
linear control under the common UE and baseline receiver using coherent peak search. SNR sweeps, a temporal nonlinear control, and receiver ablations
assess noise sensitivity, spectral increments, sequential recovery,
and learned local correction.
\end{enumerate}

Sections~\ref{sec:system_model} and~\ref{sec:building_blocks}
present the system model and mathematical foundations.
Section~\ref{sec:proposed_method} develops the framework;
Sections~\ref{sec:experiments} and~\ref{sec:conclusion}
report the experiments and conclusions.

\textit{Notation:}
Superscripts $(\cdot)^T$, $(\cdot)^H$, and $(\cdot)^*$ denote
transpose, Hermitian transpose, and complex conjugation.
The operators $\Re$ and $\Im$ extract real and imaginary parts;
$\|\cdot\|_2$ and $\|\cdot\|_{\mathrm F}$ are Euclidean and Frobenius norms.
For feature tensors, $[\mathbf A;\mathbf B]$ concatenates channels
and $\odot$ denotes elementwise multiplication.
The operators $\operatorname{vec}$, $\operatorname{reshape}$ and
$\operatorname{diag}$ vectorize, reshape without reordering and form
diagonal matrices; $\mathbf I_n$ is the identity.
Scalar functions act elementwise:
$\operatorname{sigmoid}(x)=(1+e^{-x})^{-1}$ and
$\operatorname{clip}_{[a,b]}(x)=\min\{\max(x,a),b\}$.
The operators $\mathcal F_{\Lsf}$ and $\mathcal F_{\Lsf}^{-1}$ denote
the unitary fast-time discrete Fourier transform (DFT) and inverse,
computed by fast Fourier transforms (FFTs).

\begin{figure}[H]
\centering
\includegraphics[width=\columnwidth]{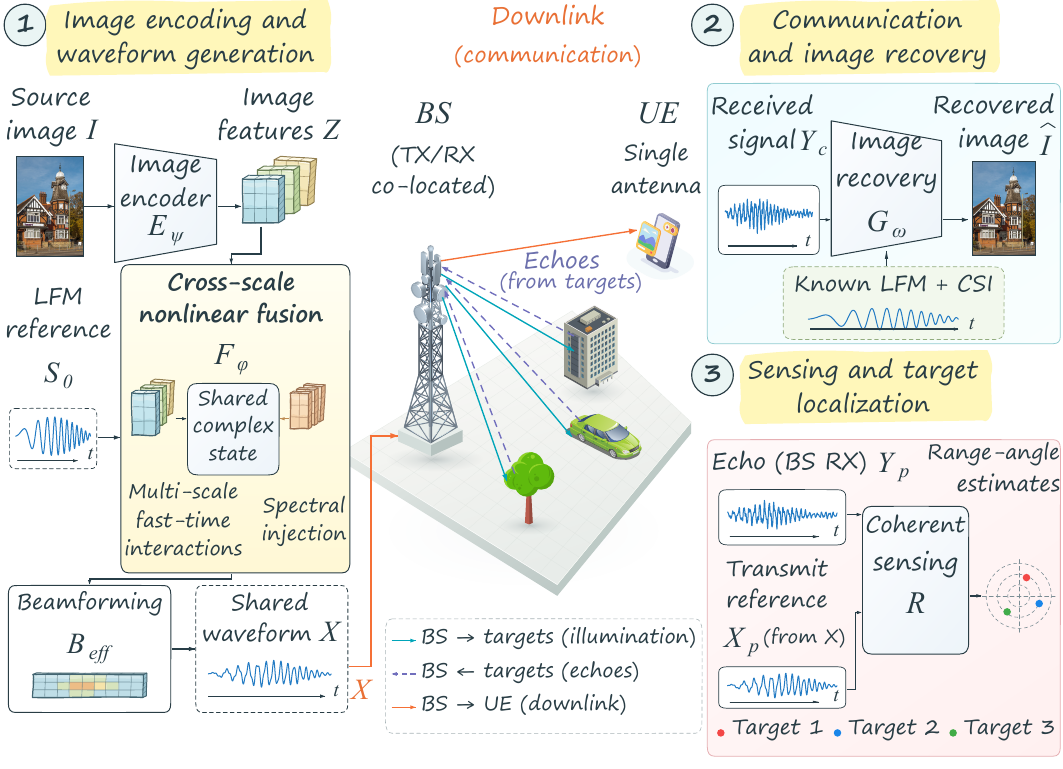}
\caption{System overview: (1) image encoding, fusion of
image and LFM features, and beamforming; (2) image recovery
using the received frame, LFM reference, and CSI;
and (3) coherent range and angle estimation
from echoes using the transmitted waveform.}
\label{fig:framework}
\end{figure}

\section{System Model and Problem Formulation}
\label{sec:system_model}

We consider a monostatic downlink ISAC system shown in
Fig.~\ref{fig:framework}, where a BS transmits an image to a UE with one antenna while localizing $\Ksf$ static point targets.
The BS employs transmit and receive uniform linear arrays
with $\Nsf_{\mathrm t}$ and $\Nsf_{\mathrm r}$ elements, respectively,
with spacing $\lambda/2$, where $\lambda$ is the carrier wavelength.

Each frame contains $\Psf$ pulses with $\Lsf$ complex fast-time
samples per pulse. The transmit frame is denoted by
$\mathbf X\in\mathbb C^{\Nsf_{\mathrm t}\times \Psf\times \Lsf}$,
with $\mathbf X_p=\mathbf X_{:,p,:}\in
\mathbb C^{\Nsf_{\mathrm t}\times \Lsf}$ representing its $p$th pulse.
The pulse and sample indices are $p=0,\ldots,\Psf-1$ and
$\ell=0,\ldots,\Lsf-1$, respectively.
The UE reconstructs the image from the full received frame.
Each sensing decision uses one echo pulse
$\mathbf Y_p\in\mathbb C^{\Nsf_{\mathrm r}\times \Lsf}$
and the corresponding realized transmit pulse $\mathbf X_p$.

\subsection{Transmit Signal Model}
\label{ssec:tx}

The BS encodes the source image $\mathbf I$ into a complex
communication representation,
\begin{equation}
\mathbf Z=E_{\boldsymbol\psi}(\mathbf I)
\in\mathbb C^{\Psf\times \Lsf},
\label{eq:encoder_map}
\end{equation}
where $E_{\boldsymbol\psi}$ is an image encoder with trainable
parameters $\boldsymbol\psi$. Its output is normalized to
$\|\mathbf Z\|_{\mathrm F}^2=2\Psf\Lsf$.
The encoder architecture is described in
Section~\ref{subsec:image_representation}.

The communication representation $\mathbf Z$ is fused with a
deterministic LFM reference $\mathbf S_0\in\mathbb C^{\Psf\times \Lsf}$.
The reference uses the same chirp in each pulse:
\begin{equation}
[\mathbf S_0]_{p,\ell}
=
\exp\!\left[
j2\pi\left(-\frac{B}{2}t_\ell+\frac{\mu}{2}t_\ell^2\right)
\right],
\label{eq:lfm}
\end{equation}
where $t_\ell=\ell/f_{\mathrm s}$, $f_{\mathrm s}$ is the
sampling rate, $B$ is the chirp bandwidth, and
$\mu=Bf_{\mathrm s}/\Lsf$ is the chirp slope.
Each reference sample has unit power,
$|[\mathbf S_0]_{p,\ell}|^2=1$.

Before fusion, $\mathbf Z$ and $\mathbf S_0$ are each expanded
into $\Msf=4$ transformed copies, one for each spatial mode.
For either source $\mathbf V\in\mathbb C^{\Psf\times \Lsf}$,
the $m$th copy is obtained by a circular shift of $m\Lsf/\Msf$
fast-time samples and a \mbox{phase rotation} set by the pulse index:
\begin{equation}
[\mathcal E_m(\mathbf V)]_{p,\ell}
=
e^{j2\pi mp/\Msf}
[\mathbf V]_{p,(\ell-m\Lsf/\Msf)\bmod \Lsf},
\label{eq:mode_expansion}
\end{equation}
where $m=0,\ldots,\Msf-1$ and $\Lsf$ is divisible by $\Msf$.
Stacking the copies gives
$\mathcal E(\mathbf V)\in\mathbb C^{\Msf\times \Psf\times \Lsf}$.
Each copy preserves the source energy, so
$\|\mathcal E(\mathbf V)\|_{\mathrm F}^2
=\Msf\|\mathbf V\|_{\mathrm F}^2$.

Transmit beamforming uses the communication CSI
$\mathbf h_{\mathrm c}\in\mathbb C^{\Nsf_{\mathrm t}}$
and a prescribed sensing covariance
$\bar{\mathbf R}_{\mathrm s}\in
\mathbb C^{\Nsf_{\mathrm t}\times \Nsf_{\mathrm t}}$.
The covariance is formed by averaging steering responses over
$[-45^\circ,45^\circ]$, independently of the realized target scene.
Let $\mathbf U_{\mathrm s}\in
\mathbb C^{\Nsf_{\mathrm t}\times(\Msf-1)}$ contain the eigenvectors
associated with the $\Msf-1$ largest eigenvalues of
$\bar{\mathbf R}_{\mathrm s}$.
The beamforming matrix
$\mathbf B_{\mathrm{eff}}\in\mathbb C^{\Nsf_{\mathrm t}\times \Msf}$
is constructed as
\begin{equation}
\mathbf w_{\mathrm c}
=
\frac{\mathbf h_{\mathrm c}^{*}}{\|\mathbf h_{\mathrm c}\|_2},
\qquad
\mathbf B_{\mathrm{eff}}
=
\operatorname{qf}
\bigl([\mathbf w_{\mathrm c},\mathbf U_{\mathrm s}]\bigr),
\label{eq:effective_basis}
\end{equation}
where $\operatorname{qf}$ returns the factor with orthonormal columns
of a reduced QR decomposition with a fixed phase convention.
We assume $\mathbf h_{\mathrm c}\neq\mathbf 0$ and
$\operatorname{rank}([\mathbf w_{\mathrm c},\mathbf U_{\mathrm s}])=\Msf$,
so that
$\mathbf B_{\mathrm{eff}}^{H}\mathbf B_{\mathrm{eff}}=\mathbf I_{\Msf}$.
The first column provides a communication beam aligned with
$\mathbf h_{\mathrm c}^{*}$, while the remaining columns
provide sensing beams in the null space of
$\mathbf h_{\mathrm c}^{T}$. The fusion mapping $F_{\boldsymbol\phi}$ jointly processes the
expanded communication and LFM inputs to generate the
$\Msf$ \mbox{mode waveforms}:
\begin{equation}
\mathbf S_{\mathrm{raw}}
=
F_{\boldsymbol\phi}
\bigl(\mathcal E(\mathbf S_0),\mathcal E(\mathbf Z);\mathbf c\bigr)
\in\mathbb C^{\Msf\times \Psf\times \Lsf},
\label{eq:nonlinear_map}
\end{equation}
where $\boldsymbol\phi$ denotes the trainable fusion parameters.
The real context vector $\mathbf c$ is constructed from
$\mathbf h_{\mathrm c}$, $\bar{\mathbf R}_{\mathrm s}$,
and descriptors of $\mathbf B_{\mathrm{eff}}$.
The context construction and waveform fusion architecture are
described in Sections~\ref{subsec:image_representation}
and~\mbox{\ref{sec:fuser}, respectively}.

Beamforming produces the unnormalized transmit pulse
\begin{equation}
\widetilde{\mathbf X}_p
=
\mathbf B_{\mathrm{eff}}\mathbf S_{\mathrm{raw},p},
\label{eq:array_mapping}
\end{equation}
where $\mathbf S_{\mathrm{raw},p}
=\mathbf S_{\mathrm{raw},:,p,:}\in\mathbb C^{\Msf\times \Lsf}$.
Let $\widetilde{\mathbf X}$ collect the $\Psf$ beamformed pulses.
For $\|\widetilde{\mathbf X}\|_{\mathrm F}>0$,
the transmitted frame is normalized as
\begin{equation}
\mathbf X
=
\alpha_{\mathrm{norm}}\widetilde{\mathbf X},
\qquad
\alpha_{\mathrm{norm}}
=
\frac{\sqrt{\Psf\Lsf}}{\|\widetilde{\mathbf X}\|_{\mathrm F}}.
\label{eq:array_waveform}
\end{equation}
The common scaling factor enforces
$\|\mathbf X\|_{\mathrm F}^2
=\sum_{p=0}^{\Psf-1}\|\mathbf X_p\|_{\mathrm F}^2=\Psf\Lsf$.
This energy constraint applies to the entire array frame,
without separate constraints on individual antennas or pulses.

\subsection{Communication Model}
\label{ssec:comm}

The channel from the BS to the UE $\mathbf h_{\mathrm c}$ is flat across frequency
and remains constant over each frame. The UE receives
\begin{equation}
y_{\mathrm c}[p,\ell]
=
[\mathbf h_{\mathrm c}^{T}\mathbf X_p]_\ell
+n_{\mathrm c}[p,\ell],
\label{eq:yc}
\end{equation}
where the additive noise samples are independent and
identically distributed (i.i.d.) as
$\mathcal{CN}(0,\sigma_{\mathrm c}^2)$ within each frame.
The received samples form
$\mathbf Y_{\mathrm c}\in\mathbb C^{\Psf\times \Lsf}$,
with $[\mathbf Y_{\mathrm c}]_{p,\ell}=y_{\mathrm c}[p,\ell]$.

Let $\mathbf b_0$ denote the first column of
$\mathbf B_{\mathrm{eff}}$ and
$g_{\mathrm c}=\mathbf h_{\mathrm c}^{T}\mathbf b_0$
the effective channel gain.
The beamforming construction 
gives
\begin{equation}
\begin{aligned}
\mathbf h_{\mathrm c}^{T}\mathbf B_{\mathrm{eff}}
&=[g_{\mathrm c},0,\ldots,0],\\
y_{\mathrm c}[p,\ell]
&=\alpha_{\mathrm{norm}}g_{\mathrm c}
[\mathbf S_{\mathrm{raw},p}]_{0,\ell}
+n_{\mathrm c}[p,\ell].
\end{aligned}
\label{eq:ue_spatial_reduction}
\end{equation}
The UE observes the mode aligned with the communication channel,
which is jointly synthesized from the image representation
and the LFM reference.
It reconstructs the image from the full received frame
$\mathbf Y_{\mathrm c}$ using the known reference
$\mathbf S_0$ and CSI $\mathbf h_{\mathrm c}$.

\subsection{Sensing Model}
\label{ssec:sens}

For $u=\sin\theta$, the transmit and receive steering vectors
$\mathbf a_{\mathrm t}(u)\in\mathbb C^{\Nsf_{\mathrm t}}$
and $\mathbf a_{\mathrm r}(u)\in\mathbb C^{\Nsf_{\mathrm r}}$
are defined by
\begin{equation}
[\mathbf a_{\mathrm t}(u)]_n=e^{j\pi nu},
\qquad
[\mathbf a_{\mathrm r}(u)]_n=e^{-j\pi nu},
\label{eq:steering_vector}
\end{equation}
where $n=0,\ldots,\Nsf_{\mathrm t}-1$ for the transmit array
and $n=0,\ldots,\Nsf_{\mathrm r}-1$ for the receive array.
The transmitted waveform in direction $u$ during pulse $p$ is
\[
q_p(u)=\mathbf a_{\mathrm t}^{T}(u)\mathbf X_p
\in\mathbb C^{1\times \Lsf}.
\]

Target $k$ has range $r_k$, angle $\theta_k$, and echo
coefficient $\alpha_k$. We define $u_k=\sin\theta_k$ and
the round trip delay $\tau_k=2f_{\mathrm s}r_k/c_0$ in
samples, where $c_0$ is the speed of light.
The target parameters remain constant over the frame,
with no Doppler progression.

We adopt an idealized echo model with circular blocks of length $\Lsf$.
Echo generation and coherent processing use periodic
extension, without modeling a cyclic prefix or a receive window for a finite pulse. The operator for fractional delay
$\mathcal D_\tau$ is defined by
\begin{equation}
\mathcal F_{\Lsf}\bigl(\mathcal D_\tau(q)\bigr)[b]
=
\mathcal F_{\Lsf}(q)[b]e^{-j2\pi f_b\tau},
\label{eq:delay}
\end{equation}
where $f_b$ is the signed frequency of bin
$b=0,\ldots,\Lsf-1$ in cycles per sample.
The received echo for pulse $p$ is
\begin{equation}
\mathbf Y_p
=
\sum_{k=1}^{\Ksf}
\alpha_k\mathbf a_{\mathrm r}(u_k)
\mathcal D_{\tau_k}\bigl(q_p(u_k)\bigr)
+\mathbf N_p,
\label{eq:ys}
\end{equation}
where $\mathbf N_p$ has independent
$\mathcal{CN}(0,\sigma_{\mathrm s}^2)$ entries.

Using the known realized transmit pulse $\mathbf X_p$
and target count $\Ksf$, the BS estimates the unknown target
coordinates from $\mathbf Y_p$. Denoting the sensing receiver by
$\mathcal R$, its output
$\widehat{\mathcal T}_p=\mathcal R(\mathbf Y_p,\mathbf X_p;\Ksf)$
contains $\Ksf$ estimated range and angle pairs.

\subsection{Problem Formulation}
\label{ssec:task_formulation}

Let $\boldsymbol\omega$ parameterize the UE demixer and image
decoder. For the transmitted waveform generated by
$E_{\boldsymbol\psi}$ and $F_{\boldsymbol\phi}$,
the reconstructed image is
\begin{equation}
\widehat{\mathbf I}=G_{\boldsymbol\omega}
(\mathbf Y_{\mathrm c},\mathbf S_0,\mathbf h_{\mathrm c}).
\label{eq:task_maps}
\end{equation}

For a selected task weight $w\in[0,1]$, waveform learning minimizes the
expected normalized communication and sensing losses:
\begin{equation}
\begin{aligned}
&\min_{\boldsymbol\psi,\boldsymbol\phi}\;
\mathbb E\!\left[
\frac{w}{n_{\mathrm c}}\mathcal L_{\mathrm c}
+\frac{1-w}{n_{\mathrm s}}\overline{\mathcal L}_{\mathrm{RA}}
\right]\\
&\text{subject to}\quad \|\mathbf X\|_{\mathrm F}^2=\Psf\Lsf.
\end{aligned}
\label{eq:task_formulation}
\end{equation}
The energy constraint follows \eqref{eq:array_waveform}.
Here, $\mathcal L_{\mathrm c}$ measures image reconstruction error, and
$\overline{\mathcal L}_{\mathrm{RA}}$ aggregates range and angle localization
losses computed from separate receiver calls for individual pulses. The
expectation is taken over image, channel, scene, and noise realizations.
The positive constants $n_{\mathrm c}$ and $n_{\mathrm s}$ place the task
losses on fixed scales shared by all compared waveform methods.

A separate waveform model is trained for each task weight $w$,
updating $\boldsymbol\psi$ and $\boldsymbol\phi$ under common,
fixed receive processing. The local sensing network is then
adapted with the transmitter fixed.

\section{Preliminaries}
\label{sec:building_blocks}

Problem~\eqref{eq:task_formulation} specifies the task objectives
and constraint on total transmit energy.
This section motivates source interaction that depends on the inputs and
introduces the complex operations and relationship
statistics used by the fusion mapping
in~\eqref{eq:nonlinear_map}.
The specific architecture is developed in
Section~\ref{sec:proposed_method}.

\subsection{From Linear Superposition to Fusion That Depends on the Inputs}
\label{subsec:fusion_motivation}

For inputs $\mathbf s=\mathcal E(\mathbf S_0)$ and
$\mathbf z=\mathcal E(\mathbf Z)$ from~\eqref{eq:mode_expansion},
linear fusion under a fixed physical context can be written
before normalization as
\begin{equation}
F_{\mathrm{lin}}(\mathbf s,\mathbf z)
=\mathbf A_{\mathrm s}\mathbf s+\mathbf A_{\mathrm c}\mathbf z,
\label{eq:linear_fusion_definition}
\end{equation}
where the operators $\mathbf A_{\mathrm s}$ and
$\mathbf A_{\mathrm c}$ are independent of the current source
features and may implement learned bandwise or mode mixing.
A perturbation $\Delta\mathbf z$ to the communication input gives
{\interdisplaylinepenalty=10000
\[
\begin{aligned}
&F_{\mathrm{lin}}(\mathbf s,\mathbf z+\Delta\mathbf z)
-F_{\mathrm{lin}}(\mathbf s,\mathbf z)\\
&\qquad=\mathbf A_{\mathrm c}\Delta\mathbf z.
\end{aligned}
\]
}
This response is independent of the current source features,
motivating interaction weights based on feature magnitudes
and relative phases, as developed in
Section~\ref{sec:proposed_method}.

\subsection{Processing of Complex Features}
\label{subsec:complex_processing}

Complex features are indexed by channel, spatial mode, pulse,
and fast-time sample. We use $\Csf$ for the channel width of
source and joint states, and $\Dsf$ for a general input width.
Complex root-mean-square (RMS) normalization $\mathcal N$
acts independently on each frame and spatial mode.
For $\mathbf V\in\mathbb C^{\Dsf\times \Msf\times \Psf\times \Lsf}$,
the channels are divided into four equal contiguous groups.
Let $\mathcal G(d)$ denote the group containing channel $d$:
\[
[\mathcal N(\mathbf V)]_{d,m,p,\ell}
=
\frac{\eta_d V_{d,m,p,\ell}}
{\sqrt{\langle|\mathbf V|^2\rangle_{\mathcal G(d),m}+10^{-6}}},
\]
where the average covers the group's channels and all
pulse and sample positions.
Each layer has learned real channel gains $\eta_d$,
initialized to one and shared over modes and positions,
with no additive bias.

Complex convolutions couple in-phase/quadrature (I/Q) components~\cite{trabelsi2018deep}.
Compact $3\times3$ kernels aggregate three neighboring pulses and
fast-time taps $(-d,0,d)$, using dilation $(1,d)$ on $(\Psf,\Lsf)$.
Varying $d$ changes scale without adding coefficients or
downsampling~\cite{yu2016multi}; circular padding wraps boundary
indices to preserve $\Psf\times \Lsf$.

At fixed $(p,\ell)$, $1\times1$ projections mix channels within
each mode; mixing across channels and modes transforms both axes using
a learned complex $\Csf\Msf\times \Csf\Msf$ matrix without bias, shared over
positions and specific to each source or joint block.

Radial activation $\mathcal A(z)=z\operatorname{sigmoid}(|z|)$
provides nonlinearity that depends on magnitude while preserving nonzero
features' phases. 

\subsection{Statistics of Magnitude and Relative Phase}
\label{subsec:relationship_statistics}

Adaptive weighting uses feature strengths
and pairwise relationships.
For nonzero complex scalars $z_1$ and $z_2$,
\[
z_1^*z_2
=
|z_1||z_2|
e^{j(\arg z_2-\arg z_1)}.
\]
The conjugate product combines magnitude information with relative
phase and is invariant to a common phase rotation of the two
features. We form real descriptors using
\[
\ell(z)=\log(1+|z|^2),\qquad
\chi(z)=[\tanh(\Re z);\tanh(\Im z)].
\]
The terms $\ell(z_1)$ and $\ell(z_2)$ describe individual
feature strengths on a compressed scale, while
$\chi(z_1^*z_2)$ provides bounded descriptors of their
relationship. The temporal and spectral blocks in
Sections~\ref{subsec:temporal_block}
and~\ref{subsec:spectral_block} use these descriptors
to construct adaptive feature weights.
Their affine sigmoid $0.25+1.5\operatorname{sigmoid}(x)$ lies
in $(0.25,1.75)$, with unit gain at zero preactivation,
allowing attenuation and amplification.

\section{Proposed Framework}
\label{sec:proposed_method}

Shared waveform design must support image recovery through
the UE channel and target localization from echoes of the
same transmission. We address this with separate source
branches and a shared complex state. The branches maintain
features from the image and LFM reference, while the shared
state accumulates their interactions according to the current
inputs. Temporal processing combines features across different
sample neighborhoods, while spectral updates couple the two
sources within frequency bands and enter the same state.
A single output head generates the mode waveforms, followed
by beamforming and normalization to the prescribed total
array frame energy. The receivers exploit the distinct
information available at the UE and BS.

\begin{figure*}[!t]
\centering
\includegraphics[width=0.96\textwidth]{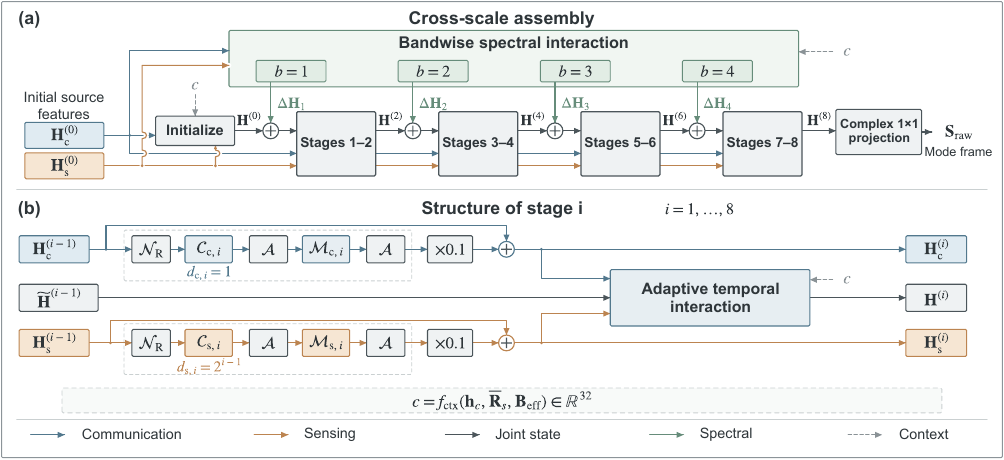}
\caption{Waveform generation across scales. (a) Joint state evolution over eight stages,
with spectral increments injected before stages 1, 3, 5, and 7
and a final projection to $\mathbf S_{\mathrm{raw}}$.
(b) Updates of the source branches and temporal interaction within one stage.}
\label{fig:overall_architecture}
\end{figure*}

\subsection{Nonlinear Waveform Fusion Across Scales}
\label{sec:fuser}
\label{sec:encoder}

\subsubsection{Image representation and feature initialization}
\label{subsec:image_representation}

The image encoder $E_{\boldsymbol\psi}$ in
\eqref{eq:encoder_map} is a convolutional network that
produces $2\Csf_z$ real feature channels at $8\times8$
spatial resolution. Before complex pairing, the output
is normalized to unit mean square over its $2\Psf\Lsf$ entries.
Denoting the normalized in-phase and quadrature components by
$E_{\boldsymbol\psi}^{\mathrm I}(\mathbf I)$
and $E_{\boldsymbol\psi}^{\mathrm Q}(\mathbf I)$, respectively,
the encoding map is
\begin{equation}
E_{\boldsymbol\psi}(\mathbf I)
=\operatorname{reshape}_{\Psf\times \Lsf}\!\bigl(
E_{\boldsymbol\psi}^{\mathrm I}(\mathbf I)
+jE_{\boldsymbol\psi}^{\mathrm Q}(\mathbf I)\bigr).
\label{eq:image_encoder}
\end{equation}
Here, $\Csf_z=64$ gives $\Csf_z\cdot8\cdot8=\Psf\Lsf=4096$
complex values. This direct mapping from images to channel symbols~\cite{bourtsoulatze2019deep} provides $\mathbf Z$
for mode expansion and generation of source features
in~\eqref{eq:initial_source_features}.

The expanded inputs $\mathbf s$ and $\mathbf z$ each contain
one complex value per mode and pulse and sample position.
Separate complex $1\times1$ convolutions
$\mathcal P_{\mathrm s}$ and $\mathcal P_{\mathrm c}$ map
these values to $\Csf=4$ feature channels:
\begin{equation}
\mathbf H_{\mathrm s}^{(0)}=\mathcal P_{\mathrm s}(\mathbf s),
\qquad
\mathbf H_{\mathrm c}^{(0)}=\mathcal P_{\mathrm c}(\mathbf z).
\label{eq:initial_source_features}
\end{equation}
Both tensors belong to
$\mathbb C^{\Csf\times \Msf\times \Psf\times \Lsf}$, with projection
weights shared over modes and pulse and sample positions.
The superscript $(0)$ denotes the initial features before
\mbox{updates of the source branches}.

The beamforming modes have different communication and sensing
roles. To account for these roles during joint synthesis,
we construct a context vector $\mathbf c\in\mathbb R^{32}$
from CSI, the prescribed sensing covariance, and beamforming descriptors for each mode.
For column $\mathbf b_m$ of $\mathbf B_{\mathrm{eff}}$, define
\[
\begin{aligned}
v_m&=\frac{\mathbf h_{\mathrm c}^{T}\mathbf b_m}
{\max(\|\mathbf h_{\mathrm c}\|_2,10^{-8})},\\
s_m&=\max\!\bigl(\Re\{\mathbf b_m^H
\bar{\mathbf R}_{\mathrm s}\mathbf b_m\},0\bigr).
\end{aligned}
\]
Here, $v_m$ describes normalized coupling to the UE channel,
whereas $s_m$ measures the energy of the prescribed sensing covariance
along beam $\mathbf b_m$.
For each mode, we use the seven descriptors
$\mathbf1_{\{m=0\}}$, $|v_m|^2$, $s_m$,
$s_m/\max(\sum_{m'}s_{m'},10^{-8})$,
$\log(s_m+10^{-8})$, $\Re v_m$, and $\Im v_m$.
Separate multilayer perceptrons embed the real/imaginary CSI,
covariance entries, and descriptors concatenated across modes;
a shared multilayer perceptron combines their outputs into
$\mathbf c$.

For $i=0,\ldots,8$, learned affine maps of $\mathbf c$ produce real
$\Csf\times \Msf$ arrays $\mathbf L_i(\mathbf c)$ and
$\boldsymbol\Phi_i(\mathbf c)$. The resulting complex gain is
\[
\mathbf K_i(\mathbf c)=\exp\!\left[
0.35\tanh\mathbf L_i(\mathbf c)
+j\tfrac{\pi}{2}\tanh\boldsymbol\Phi_i(\mathbf c)\right].
\]
The gain is broadcast over $(\Psf,\Lsf)$, with magnitude in
$(e^{-0.35},e^{0.35})$ and phase in $(-\pi/2,\pi/2)$.
For $i=0$, it conditions the initial joint state:
\begin{equation}
\mathbf H^{(0)}=\mathbf K_0(\mathbf c)\odot
\mathcal W_0[\mathbf H_{\mathrm s}^{(0)};\mathbf H_{\mathrm c}^{(0)}],
\label{eq:initial_joint_state}
\end{equation}
where $\mathcal W_0$ is a complex $1\times1$ projection
($2\Csf\!\to\!\Csf$ channels).

\subsubsection{Bandwise spectral interaction}
\label{subsec:spectral_block}

The spectral branch makes source interaction explicitly
frequency selective. Channel projections specific to each band and
mode mixing allow different spectral regions to use different combinations of sources and modes, while weights derived from the current source spectra adapt these contributions.
Both sources participate within every processing band.

Conditioned on $\mathbf c$, the branch maps the initial
features $\mathbf H_{\mathrm s}^{(0)}$ and
$\mathbf H_{\mathrm c}^{(0)}$ to four increments in the time domain
$\Delta\mathbf H_b\in
\mathbb C^{\Csf\times\Msf\times\Psf\times\Lsf}$,
$b=1,\ldots,4$.
These increments are computed once per frame, providing
spectral contributions that depend on the sources that subsequent
temporal updates incorporate into the joint state.

\begin{figure}[!t]
\centering
\includegraphics[width=0.90\columnwidth]{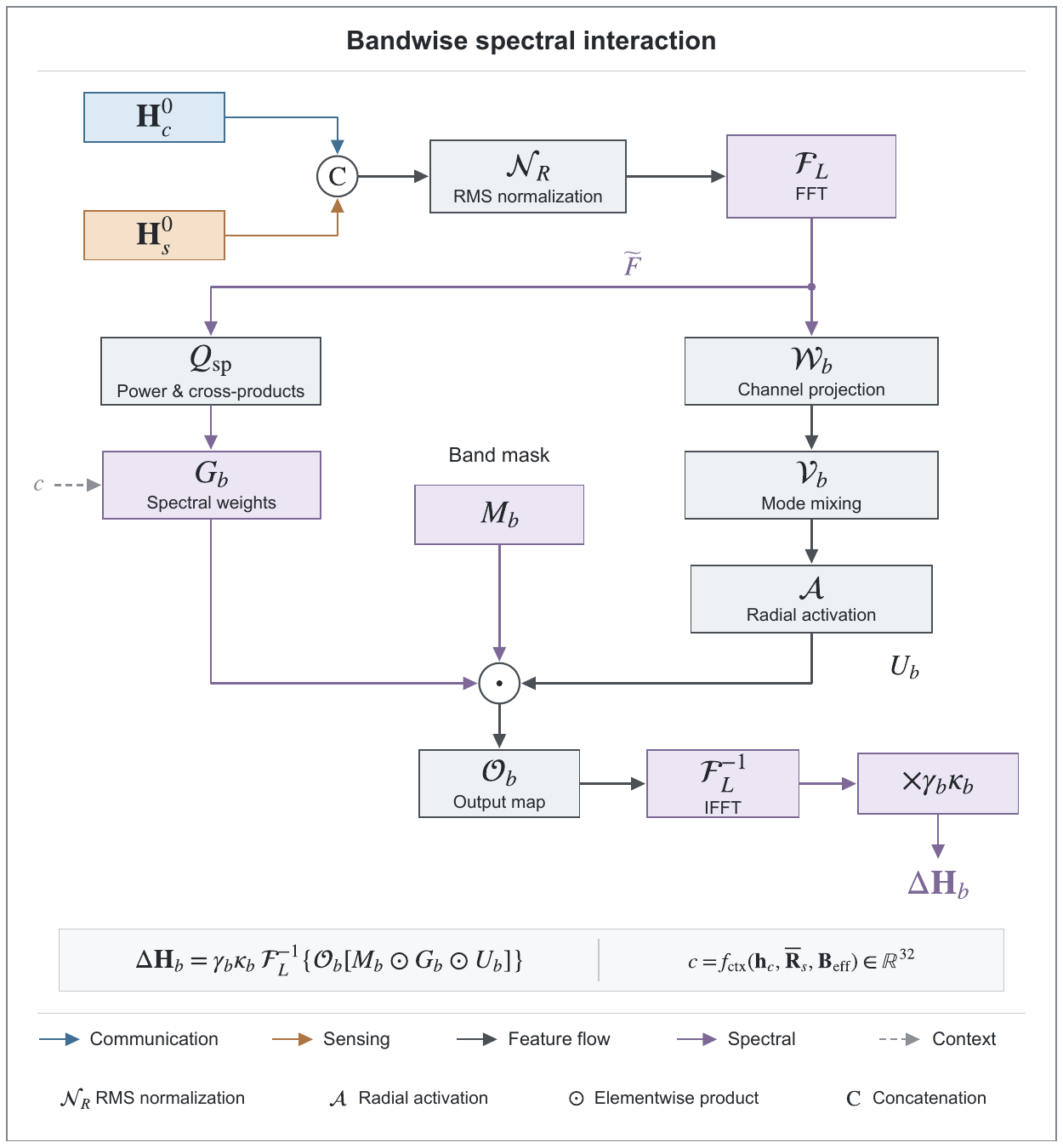}
\caption{Spectral interaction within band $b$.
Initial source spectra provide candidate features and adaptive weights.
Mask $M_b$, output projection, inverse DFT, and scaling by
$\gamma_b\kappa_b$ produce the increment $\Delta\mathbf H_b$.
Here, $\mathcal N_{\mathrm R}=\mathcal N$.}
\label{fig:spectral_interaction}
\end{figure}

The normalized source spectra are
\[
\begin{aligned}
\widehat{\mathbf F}
&=\mathcal F_{\Lsf}\!\left\{\mathcal N
([\mathbf H_{\mathrm s}^{(0)};\mathbf H_{\mathrm c}^{(0)}])\right\}\\
&=[\mathbf F_{\mathrm s};\mathbf F_{\mathrm c}].
\end{aligned}
\]
For band $b$, let $M_b\in\{0,1\}^{\Lsf}$ be a fixed binary
mask, equal to one at its FFT bins and zero elsewhere.
The four bands in absolute frequency normalized by the Nyquist frequency are
$[0,1/8]$, $(1/8,1/4]$, $(1/4,1/2]$, and $(1/2,1]$.
Each mask is broadcast over channels, spatial modes, and pulses.
For each band, a complex
projection $\mathbf W_b$ maps $2\Csf$ source channels to $\Csf$,
followed by mode mixing $\mathbf V_b$:
\begin{equation}
\mathbf T_b=\mathbf V_b\mathbf W_b\widehat{\mathbf F},
\qquad b=1,\ldots,4.
\label{eq:spectral_projection}
\end{equation}
The map $\mathbf V_b$ mixes spatial modes separately
within each feature channel. Each band and channel
has its own complex $\Msf\times \Msf$ matrix, shared over pulses
and \mbox{frequency bins}. Using $\ell$ and $\chi$ from Section~\ref{subsec:relationship_statistics},
the spectral relationship tensor is
\[
\begin{aligned}
\mathbf Q_{\mathrm{sp}}=[&\ell(\mathbf F_{\mathrm s});
\ell(\mathbf F_{\mathrm c});\\
&\chi(\mathbf F_{\mathrm s}^{*}\odot\mathbf F_{\mathrm c})]
\in\mathbb R^{4\Csf\times \Msf\times \Psf\times \Lsf}.
\end{aligned}
\]
A real $1\times1$ convolution $\mathcal L_b$ maps these
$4\Csf$ channels to $\Csf$. A learned real affine map with
$\mathbf A_b^{\mathrm{sp}}\in\mathbb R^{\Csf\Msf\times32}$ and
$\mathbf b_b^{\mathrm{sp}}\in\mathbb R^{\Csf\Msf}$ supplies the context offset:

\[
\begin{aligned}
\mathbf G_b={}&0.25+1.5\operatorname{sigmoid}\!\bigl[
\mathcal L_b(\mathbf Q_{\mathrm{sp}})\\
&+\operatorname{reshape}_{\Csf\times \Msf}
(\mathbf A_b^{\mathrm{sp}}\mathbf c+\mathbf b_b^{\mathrm{sp}})
\bigr].
\end{aligned}
\]
The convolution is shared across modes; context offsets are
broadcast over pulses and frequency bins. The weights retain
shape $\Csf\times \Msf\times \Psf\times \Lsf$ without pooling.

The activated features $\mathbf U_b=\mathcal A(\mathbf T_b)$
are weighted, restricted to their band, and processed by the complex
projection $\mathbf O_b\in\mathbb C^{\Csf\times\Csf}$,
shared over modes, pulses, and frequency bins. The increment in the time domain is
\begin{equation}
\Delta\mathbf H_b=\gamma_b\kappa_b\mathcal F_{\Lsf}^{-1}
\!\left\{\mathbf O_b
[M_b\odot\mathbf G_b\odot\mathbf U_b]\right\}.
\label{eq:spectral_update}
\end{equation}
The scales are parameterized by
\begin{equation}
\begin{aligned}
\kappa_b&=0.35\operatorname{sigmoid}(\kappa_b^{\mathrm{raw}}),\\
\gamma_b&=\tfrac12\tanh(4g_b),\qquad
g_b\big|_{\mathrm{init}}=0.
\end{aligned}
\label{eq:spectral_release}
\end{equation}
The learned scalars are specific to each band and independent
of $\mathbf c$, with $\kappa_b\in(0,0.35)$.
Zero initialization of $\gamma_b$ suppresses the increment
at the start of training; $\gamma_b\kappa_b$ subsequently
controls its magnitude and sign. The initialization of
$\kappa_b$ is specified in Section~\ref{subsec:training_evaluation}.

\subsubsection{Adaptive temporal interaction}
\label{subsec:temporal_block}

At stage $i=1,\ldots,8$, the two source branches first
update their features. For $q\in\{\mathrm s,\mathrm c\}$,
the residual block~\cite{he2016deep} gives
\begin{equation}
\begin{aligned}
\mathbf V_{q,i}
&=\mathcal C_{q,i}\bigl(\mathcal N(\mathbf H_q^{(i-1)})\bigr),\\
\mathbf H_q^{(i)}
&=\mathbf H_q^{(i-1)}
+0.1\,\mathcal A\bigl(\mathcal M_{q,i}(\mathcal A(\mathbf V_{q,i}))\bigr),
\end{aligned}
\label{eq:source_feature_update}
\end{equation}
where $\mathcal C_{q,i}$ is a complex $3\times3$ convolution
with fast-time dilation $d_{q,i}$, and $\mathcal M_{q,i}$
performs the joint mixing of channels and modes defined in
Section~\ref{subsec:complex_processing}. The source and joint convolutions use the fast-time dilation sequences
\begin{equation}
\begin{aligned}
\mathbf d_{\mathrm s}&=(1,2,4,8,16,32,64,128),\\
\mathbf d_{\mathrm c}&=(1,1,1,1,1,1,1,1),\\
\mathbf d_{\mathrm j}&=(1,2,4,8,1,2,4,8).
\end{aligned}
\label{eq:dilations}
\end{equation}
The sensing branch processes features derived from the LFM reference at
progressively wider sample spacings, whereas the communication
branch uses undilated convolutions on neighboring samples.
The joint branch repeats $(1,2,4,8)$ to combine these
representations over multiple fast-time neighborhoods.

The previous joint state $\mathbf H^{(i-1)}$ receives
the spectral increment assigned to stage $i$:
\begin{equation}
\widetilde{\mathbf H}^{(i-1)}=
\begin{cases}
\mathbf H^{(i-1)}+\Delta\mathbf H_{(i+1)/2},
&i\in\{1,3,5,7\},\\
\mathbf H^{(i-1)},&i\in\{2,4,6,8\}.
\end{cases}
\label{eq:spectral_injection}
\end{equation}
The temporal block combines $\widetilde{\mathbf H}^{(i-1)}$
with the updated source features $\mathbf H_{\mathrm s}^{(i)}$
and $\mathbf H_{\mathrm c}^{(i)}$.
Two paths process their normalized concatenation: one
generates complex candidate features, while the other
derives real weights from feature magnitudes and
pairwise phase relationships.
The normalized input is
\[
\begin{aligned}
\mathbf F_i
&=\mathcal N\bigl([\widetilde{\mathbf H}^{(i-1)};
\mathbf H_{\mathrm s}^{(i)};\mathbf H_{\mathrm c}^{(i)}]\bigr)\\
&=[\mathbf U_h;\mathbf U_s;\mathbf U_c],
\end{aligned}
\]
where $\mathbf F_i$ contains $3\Csf$ channels and
$\mathbf U_h$, $\mathbf U_s$, and $\mathbf U_c$
each contain $\Csf$ channels.

\begin{figure}[t]
\centering
\includegraphics[width=0.90\columnwidth]{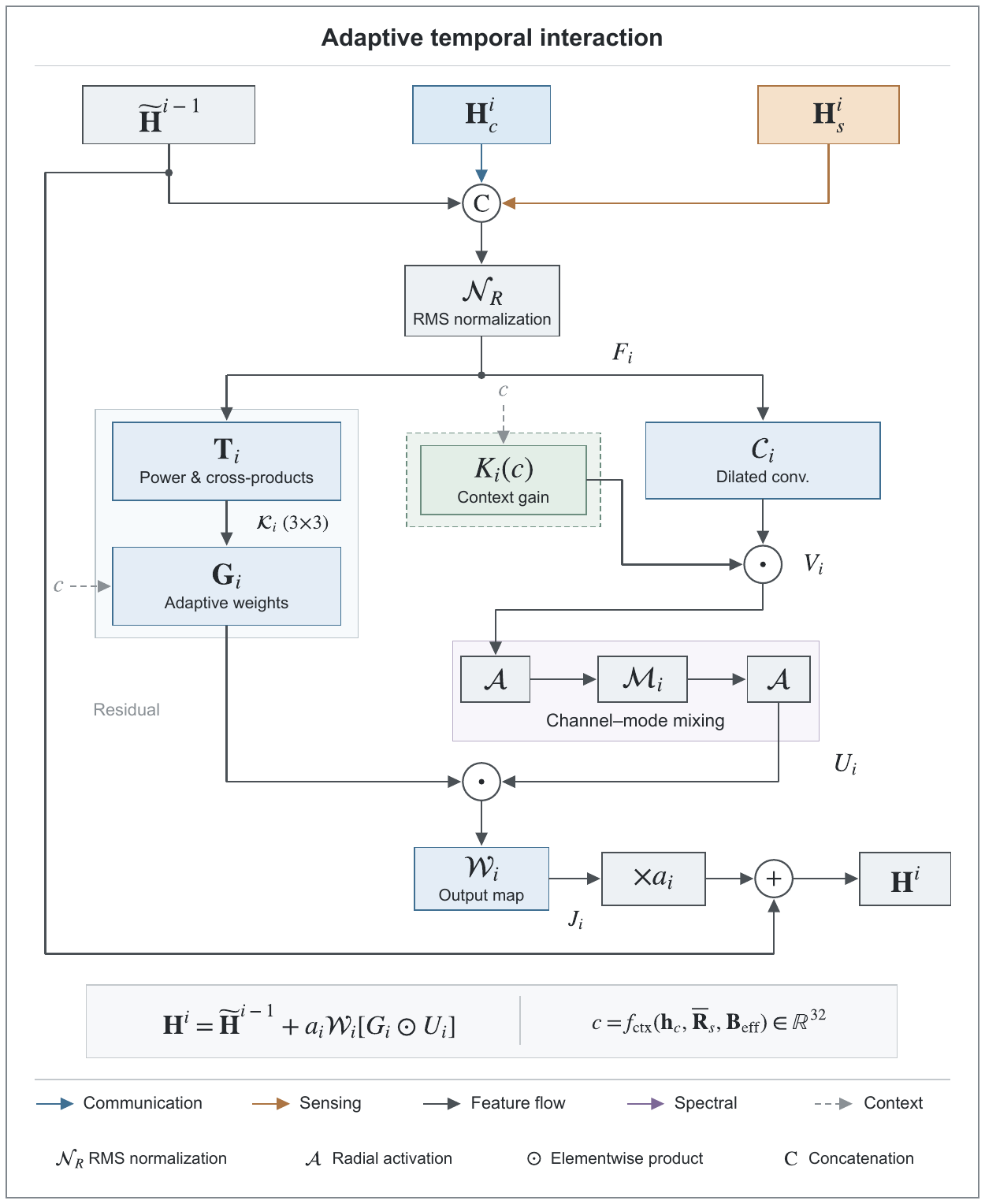}
\caption{Temporal interaction at stage $i$.
Normalized joint and source features generate candidates modulated by context and weights that vary across positions. Output projection and
residual scaling update the joint state.
Here, $\mathcal N_{\mathrm R}=\mathcal N$.}
\label{fig:temporal_interaction}
\end{figure}

The weighting path uses relationships between the two sources and between each source and the joint state. The conjugate product
$\mathbf U_s^*\odot\mathbf U_c$ describes the current relationship between the image and reference, while
$\mathbf U_h^*\odot\mathbf U_s$ and
$\mathbf U_h^*\odot\mathbf U_c$ relate each source to
the joint representation being updated.
Together with feature powers, these terms make the weights
depend on both source content and the current synthesis state.
Using $\ell$ and $\chi$ from
Section~\ref{subsec:relationship_statistics}, define
\[
\begin{aligned}
\mathbf T_i=[&\ell(\mathbf U_h);\ell(\mathbf U_s);
\ell(\mathbf U_c);\chi(\mathbf U_s^*\odot\mathbf U_c);\\
&\chi(\mathbf U_h^*\odot\mathbf U_s);
\chi(\mathbf U_h^*\odot\mathbf U_c)]
\in\mathbb R^{9\Csf\times \Msf\times \Psf\times \Lsf}.
\end{aligned}
\]

A real $3\times3$ circular convolution $\mathcal K_i$,
shared across modes, maps $9\Csf$ channels to $\Csf$.
The affine parameters specific to each stage $\mathbf A_i,\mathbf b_i$
have the same dimensions as their spectral counterparts.
The adaptive weights are
\[
\begin{aligned}
\mathbf G_i={}&0.25+1.5\operatorname{sigmoid}\!\bigl[
\mathcal K_i(\mathbf T_i)\\
&+\operatorname{reshape}_{\Csf\times \Msf}
(\mathbf A_i\mathbf c+\mathbf b_i)\bigr].
\end{aligned}
\]
The context offsets are broadcast over $(\Psf,\Lsf)$.

The candidate path combines temporal convolution, context modulation,
radial activation and joint mixing of channels and modes, followed by
adaptive weighting and projection:
\[
\begin{aligned}
\mathbf V_i&=\mathbf K_i(\mathbf c)\odot\mathcal C_i(\mathbf F_i),\\
\mathbf U_i&=\mathcal A\bigl(\mathcal M_i(\mathcal A(\mathbf V_i))\bigr),\\
\mathcal J_i&=\mathcal W_i(\mathbf G_i\odot\mathbf U_i).
\end{aligned}
\]
Here, $\mathcal C_i$ maps $3\Csf$ channels to $\Csf$ with
fast-time dilation $d_{\mathrm j,i}$,
$\mathcal M_i$ performs joint mixing of channels and modes, and
$\mathcal W_i$ is a complex $1\times1$ output projection.
The complex gain $\mathbf K_i(\mathbf c)$ uses the context to modulate
amplitude and phase, with the same gain at all pulse and sample positions.
The real weights $\mathbf G_i$ instead vary with the current
feature relationships at each position.
Thus, even at fixed physical context, changing the source
features can change the weights applied to the joint candidate
$\mathbf U_i$, rather than only changing its values.

With residual scale
$a_i=0.5\operatorname{sigmoid}(a_i^{\mathrm{raw}})\in(0,0.5)$,
the block output is
\begin{equation}
\mathbf H^{(i)}=\widetilde{\mathbf H}^{(i-1)}+a_i\mathcal J_i.
\label{eq:joint_update}
\end{equation}
The scalar $a_i^{\mathrm{raw}}$ is learned, with initialization
specified in Section~\ref{subsec:training_evaluation}.
Figure~\ref{fig:temporal_interaction} summarizes the complete block.

\subsubsection{Assembly across scales and waveform generation}
\label{subsec:cross_scale_assembly}

The generator in Fig.~\ref{fig:overall_architecture}
combines the source updates in~\eqref{eq:source_feature_update},
spectral injection in~\eqref{eq:spectral_injection},
and temporal interaction in~\eqref{eq:joint_update}
over eight stages.
Each spectral injection is followed by two temporal blocks
before the next injection or the output head.

The source branches supply current image and LFM features, while the joint state retains earlier interactions.
Spectral increments modify this state before candidate generation
and adaptive weighting, influencing subsequent temporal interactions. After stage 8, a single complex $1\times1$ output head maps
$\mathbf H^{(8)}$ to $\mathbf S_{\mathrm{raw}}$.
Beamforming and the common normalization of frame energy
in~\eqref{eq:array_mapping}--\eqref{eq:array_waveform}
then produce the transmitted array frame $\mathbf X$.

\subsection{Reception Using Available Waveform Information}
\label{sec:knowledge_rx}

\subsubsection{UE: Demixing Using the Reference and CSI}
\label{sec:comm_rx}

For pulse $p$, the real and imaginary parts of
$y_{\mathrm c}[p,:]$ and $[\mathbf S_0]_{p,:}$ form four channels for a dilated Conv1D demixer with adaptive feature weighting. A separate encoder maps the real and
imaginary CSI to affine modulation of each feature~\cite{perez2018film}.
An input convolution forms hidden features $\mathbf h_0$. At layer $i$,
CSI supplies gains $\boldsymbol\gamma_i$ and offsets $\boldsymbol\beta_i$
for group normalization $\operatorname{GN}$ with one group~\cite{wu2018group}.
A dilated convolution of the conditioned features $\widetilde{\mathbf h}_i$
produces content and weighting fields $\mathbf f_i,\mathbf g_i$. The update is
\begin{equation}
\begin{aligned}
\widetilde{\mathbf h}_i&=(1+\boldsymbol\gamma_i)\odot
\operatorname{GN}(\mathbf h_i)+\boldsymbol\beta_i,\\
\mathbf h_{i+1}&=\mathbf h_i+
\tanh(\mathbf f_i)\odot\operatorname{sigmoid}(\mathbf g_i).
\end{aligned}
\label{eq:learned_demix}
\end{equation}

The gains and offsets are broadcast over fast time. An output projection
produces two channels per pulse; stacking
and reshaping gives a $2\Csf_z\times8\times8$ frame representation for a
shared refinement network and image generator. These receiver features are trained through the image reconstruction objective.

\subsubsection{BS: Coherent Sensing Using the Realized Waveform}
\label{sec:sens_rx}

The baseline digital signal processing (DSP) and proposed sequential
receivers use dictionaries
constructed from the realized pulse $\mathbf X_p$.
The baseline selects $\Ksf$ targets from one map of coherent scores;
Algorithm~\ref{alg:sequential_sensing} gives the proposed recovery procedure.

\textit{Coherent search using the realized waveform:}
For coordinates $\mathbf q=(\tau,u)$ representing delay and sine of angle, the echo atom is
\begin{equation}
\mathbf a_p(\tau,u)=\operatorname{vec}\!\left[
\mathbf a_{\mathrm r}(u)\mathcal D_{\tau}
\bigl(\mathbf a_{\mathrm t}^{T}(u)\mathbf X_p\bigr)\right].
\label{eq:echo_atom}
\end{equation}
With $\mathbf d_p(\tau,u)=\mathbf a_p(\tau,u)/
\|\mathbf a_p(\tau,u)\|_2$ and
$\bar{\mathbf y}_p=\operatorname{vec}(\mathbf Y_p)/\sigma_{\mathrm s}$,
the coherent score is
\begin{equation}
\Lambda_p(\tau,u)=
|\mathbf d_p(\tau,u)^H\bar{\mathbf y}_p|^2.
\label{eq:mf}
\end{equation}
The physically masked grid $\mathcal Q$ has unit delay spacing
and $\Delta u=2/64$; its bounds and masking are specified
in Section~\ref{sec:exp_setup}. The baseline uses coherent matched filtering~\cite{richards2005fundamentals},
peak search, hard nonmaximum suppression within six angle indices and three delay indices of each selected grid peak, and
parabolic interpolation between bins.

Sequential recovery follows greedy detection and continuous
refinement~\cite{mamandipoor2016newtonized}. For active atoms of unit norm collected in $\mathbf B$ and ridge parameter $\epsilon=10^{-6}$,
fit amplitudes and form the residual as
\begin{equation}
\widehat{\mathbf b}=(\mathbf B^H\mathbf B+\epsilon\mathbf I)^{-1}
\mathbf B^H\bar{\mathbf y}_p,\qquad
\mathbf r=\bar{\mathbf y}_p-\mathbf B\widehat{\mathbf b}.
\label{eq:amplitude_fit}
\end{equation}
Writing $\mathbf d_q=\mathbf d_p(q)$, the next candidate maximizes
\begin{equation}
\frac{|\mathbf d_q^H\mathbf r|^2}{\max(e_q,\epsilon_e)},\qquad
e_q=1-\mathbf d_q^H\mathbf B
(\mathbf B^H\mathbf B+\epsilon\mathbf I)^{-1}\mathbf B^H\mathbf d_q.
\label{eq:sequential_score}
\end{equation}
For $\epsilon=0$ and $\mathbf B$ of full column rank, $e_q$ is the
candidate energy orthogonal to the active span.
Candidates with $e_q\leq\epsilon_e=10^{-5}$, atom norm at most
$10^{-7}$ (invisible responses), or invalid physical coordinates are excluded.

\textit{Local learned correction:}
A shared convolutional neural network (CNN) $C_{\boldsymbol\vartheta}$
maps local residual correlations and point-spread function (PSF) features of the waveform to two real outputs.
For selected point $q_0$ and its $5\times5$ grid of neighbors $q_j$ in angle and delay, define
$\chi_j=\mathbf d_p(q_j)^H\mathbf r$ and
$\chi_0=\mathbf d_p(q_0)^H\mathbf r$.
With numerical floors $\epsilon_z=10^{-6}$ and
$\epsilon_a=10^{-7}$, the correlation, PSF, and logarithmic power features are
\begin{align*}
z_j&=\frac{\chi_j\chi_0^*}{|\chi_0|^2+\epsilon_z},\\
s_j&=\mathbf d_p(q_j)^H
\frac{\mathbf a_p(q_0)}{\max(\|\mathbf a_p(q_0)\|_2,\epsilon_a)},\\
\ell_j&=\log\!\left(
1+\frac{|\chi_j|^2}{\max(e_{q_j},\epsilon_e)}\right).
\end{align*}
The six channels of $\mathbf P_q\in\mathbb R^{6\times5\times5}$ are
\[
v_j(\Re z_j,\Im z_j,\Re s_j,\Im s_j,\ell_j,1),
\]
where $v_j=1$ for neighbors inside $\mathcal Q$ and zero otherwise.
Relative correlations remove the center phase; PSF features
describe the local dictionary response.

Parabolic fits to three points of the log score in~\eqref{eq:sequential_score}
yield $\boldsymbol\delta_{\mathrm{quad}}$ in grid units along both axes,
clipped to $[-1/2,1/2]$ and zero for nonconcave fits.
The CNN then corrects the initializer:
\begin{equation}
\begin{aligned}
\mathbf q_{\mathrm{init}}&=\mathbf q_{\mathrm{grid}}+
\operatorname{diag}(1,\Delta u)\boldsymbol\delta,\\
\boldsymbol\delta&=\operatorname{clip}_{[-1/2,1/2]}
\!\left(\boldsymbol\delta_{\mathrm{quad}}
+\tfrac14\tanh C_{\boldsymbol\vartheta}(\mathbf P_q)\right).
\end{aligned}
\label{eq:neural_correction}
\end{equation}
Here, $\mathbf q_{\mathrm{grid}}=q_0$.

\textit{Physical refinement:}
For $j$ active targets, variable projection~\cite{golub1973differentiation}
jointly refines $\boldsymbol\xi=(\tau_1,u_1,\ldots,\tau_j,u_j)^T$ to reduce
$R(\boldsymbol\xi)=\|\bar{\mathbf y}_p-
\mathbf B(\boldsymbol\xi)\widehat{\mathbf b}(\boldsymbol\xi)\|_2^2$,
with amplitudes from~\eqref{eq:amplitude_fit}.
Let
$\widehat\beta_k=\widehat b_k/\|\mathbf a_p(\tau_k,u_k)\|_2$,
and let $\mathbf D_j=\operatorname{diag}(1,\Delta u,\ldots,1,\Delta u)$
convert grid steps to coordinate units.
The approximation to the projected Jacobian is
\begin{equation}
\begin{aligned}
\mathbf J_0&=\bigl[\widehat\beta_1\partial_{\tau_1}\mathbf a_p,\,
\widehat\beta_1\partial_{u_1}\mathbf a_p,\,\ldots,\,
\widehat\beta_j\partial_{u_j}\mathbf a_p\bigr]\mathbf D_j,\\
\mathbf J&=\mathbf J_0-\mathbf B
(\mathbf B^H\mathbf B+\epsilon\mathbf I)^{-1}\mathbf B^H\mathbf J_0.
\end{aligned}
\label{eq:projected_jacobian}
\end{equation}
For the unnormalized atoms at the current coordinates,
delay differentiation multiplies the delayed spectrum by $-j2\pi f_b$;
differentiation with respect to the sine of the angle acts on both steering vectors.
The approximation omits the full derivative of amplitude refitting,
and the regularized projection is not orthogonal.

Let $\mathbf H=\Re(\mathbf J^H\mathbf J)$,
$\mathbf g=\Re(\mathbf J^H\mathbf r)$, and
$\boldsymbol\Lambda$ be diagonal with
$\Lambda_{ii}=10^{-3}\max(H_{ii},10^{-5})+10^{-6}$.
A damped step with limits of half a grid cell gives
\begin{equation}
\boldsymbol\xi'=\Pi_{\Omega}\!\left[
\boldsymbol\xi+\mathbf D_j\operatorname{clip}_{[-1/2,1/2]}
\bigl((\mathbf H+\boldsymbol\Lambda)^{-1}\mathbf g\bigr)\right],
\label{eq:physical_update}
\end{equation}
where $\Omega$ is the range and angle support in Table~\ref{tab:params},
converted to coordinates for delay and sine of angle for all active targets;
$\Pi_{\Omega}$ clips to this region.
Algorithm~\ref{alg:sequential_sensing} tests acceptance using $R$.

\begin{algorithm}[!htbp]
\caption{Sequential localization using the realized waveform}
\label{alg:sequential_sensing}
\small
\begin{algorithmic}[1]

\Require Echo $\mathbf Y_p$, realized pulse $\mathbf X_p$,
target count $\Ksf$, noise scale $\sigma_{\mathrm s}$,
masked grid $\mathcal Q$, and trained CNN
$C_{\boldsymbol\vartheta}$.

\Ensure Estimated range and angle set $\widehat{\mathcal T}_p$.

\State $\bar{\mathbf y}_p\gets
\operatorname{vec}(\mathbf Y_p)/\sigma_{\mathrm s}$.

\State Normalize visible atoms from~\eqref{eq:echo_atom} to form
$\mathbf d_p(q)$ on $\mathcal Q$.

\State Initialize $\boldsymbol\xi\gets()$,
$\mathbf B\gets[\,]$, and
$\mathbf r\gets\bar{\mathbf y}_p$.

\For{$j=1,\ldots,\Ksf$}

    \State Evaluate $e_q$ and select $q_0$
    by~\eqref{eq:sequential_score} among valid candidates.

    \State Compute $\boldsymbol\delta_{\mathrm{quad}}$
    and form the local feature patch $\mathbf P_q$
    at $q_0$.

    \State Obtain $\mathbf q_{\mathrm{init}}$
    using~\eqref{eq:neural_correction}.

    \State Append $\mathbf q_{\mathrm{init}}$ to
    $\boldsymbol\xi$ and rebuild $\mathbf B(\boldsymbol\xi)$.

    \State Refit $\widehat{\mathbf b}$ and update
    $\mathbf r$ by~\eqref{eq:amplitude_fit}.

    \State Form $\mathbf J$
    by~\eqref{eq:projected_jacobian} and propose
    $\boldsymbol\xi'$ by~\eqref{eq:physical_update}.

    \State Set $\mathbf B'\gets\mathbf B(\boldsymbol\xi')$
    and refit $\widehat{\mathbf b}'$
    by~\eqref{eq:amplitude_fit}.

    \If{$R(\boldsymbol\xi')\leq R(\boldsymbol\xi)$}
        \State $(\boldsymbol\xi,\mathbf B,\widehat{\mathbf b})
        \gets
        (\boldsymbol\xi',\mathbf B',\widehat{\mathbf b}')$.
    \EndIf

    \State $\mathbf r\gets
    \bar{\mathbf y}_p-\mathbf B\widehat{\mathbf b}$.

\EndFor

\State \Return
$\widehat{\mathcal T}_p=
\{(c_0\tau_k/(2f_{\mathrm s}),\arcsin u_k)\}_{k=1}^{\Ksf}$.

\end{algorithmic}
\end{algorithm}

\subsection{Training for Image Reconstruction and Localization}
\label{sec:training}

After reference pretraining, waveform learning updates the
image encoder and fusion operator while keeping the UE
receiver and baseline sensing DSP fixed.
The trained transmitters are then held fixed, and the local
correction network is adapted within the sequential sensing
receiver. The communication loss combines mean squared
error (MSE) and structural similarity
(SSIM)~\cite{wang2004image}:
\begin{equation}
\mathcal L_{\mathrm c}=
\operatorname{MSE}(\mathbf I,\widehat{\mathbf I})
+0.08\,[1-\operatorname{SSIM}(\mathbf I,\widehat{\mathbf I})].
\label{eq:Lcomm}
\end{equation}

Let $\mathcal T=\{(r_k,\theta_k)\}_{k=1}^{\Ksf}$ contain the true coordinates and $(\hat r_k,\hat\theta_k)$ the recovered coordinates.
Because targets are unordered, a one-to-one assignment
$\pi^*$ pairs estimates with targets using the cost
$|\Delta r|/(5\,\mathrm m)+|\Delta\theta|/(10^\circ)$.
Fixed scales $r_{\mathrm{ref}}$ and $\theta_{\mathrm{ref}}$ normalize
range and angle errors. The loss over assigned pairs is
\begin{equation}
\mathcal L_{\mathrm{RA}}=
\frac{1}{2\Ksf}\sum_{k=1}^{\Ksf}
\left(\frac{|\hat r_k-r_{\pi^*(k)}|}{r_{\mathrm{ref}}}
+\frac{|\hat\theta_k-\theta_{\pi^*(k)}|}{\theta_{\mathrm{ref}}}\right).
\label{eq:Lsens}
\end{equation}

For $\Qsf$ sampled pulse positions collected in $\mathcal P_{\Qsf}$,
the aggregate sensing loss is
\begin{equation}
\overline{\mathcal L}_{\mathrm{RA}}=
\frac{1}{\Qsf}\sum_{p\in\mathcal P_{\Qsf}}
\mathcal L_{\mathrm{RA}}\bigl(
\mathcal R(\mathbf Y_p,\mathbf X_p;\Ksf),\mathcal T\bigr).
\label{eq:pulse_loss}
\end{equation}

Uniformly sampling $\Qsf$ distinct pulse positions and averaging their
separate receiver losses estimates the loss for one pulse averaged over the frame,
with $\Psf_{\mathrm{obs}}=1$ per call.

\subsubsection{Waveform optimization}
For~\eqref{eq:task_formulation}, the forward loss uses DSP hard argmax,
suppression and interpolation between bins; a sequential decoder using soft peak estimates supplies the straight-through
backward surrogate~\cite{bengio2013estimating}.

At selection step $k$, a masked softmax of the log coherent scores at
temperature $T$ yields probabilities $p_j^{(k)}$ on the valid grid. The
surrogate coordinate is their weighted mean,
$\widehat{\mathbf t}_{\mathrm{soft},k}=\sum_jp_j^{(k)}(r_j,\theta_j)$,
where $(r_j,\theta_j)$ denotes the physical coordinates of grid point $j$.
The detached hard maximum determines the next exclusion region, so hard
and soft predictions correspond by selection step.

For each target, denote the DSP coordinate by $\widehat{\mathbf t}_{\mathrm{hard}}$
and its soft surrogate by $\widehat{\mathbf t}_{\mathrm{soft}}$.
With $\operatorname{sg}$ preserving values but stopping gradients, use
\begin{equation}
\widehat{\mathbf t}_{\mathrm{ST}}
=\operatorname{sg}\!\left(\widehat{\mathbf t}_{\mathrm{hard}}
-\widehat{\mathbf t}_{\mathrm{soft}}\right)
+\widehat{\mathbf t}_{\mathrm{soft}}.
\label{eq:ste}
\end{equation}

Discrete selections and assignments are detached. The biased surrogate
retains gradients through the physical channel and sensing reference determined by the waveform to the encoder and fusion operator.

\subsubsection{Adaptation of the Sensing Receiver}
Let $\mathcal R_{\boldsymbol\vartheta}$ denote sequential recovery with
parameters of the local network $\boldsymbol\vartheta$. For given transmit
waveforms, its adaptation minimizes normalized localization loss on cached
labeled pulses:
\begin{equation}
\min_{\boldsymbol\vartheta}\;\frac{1}{n_{\mathrm s}}
\mathbb E_{(\mathbf X_p,\mathbf Y_p,\mathcal T,\Ksf)\sim\mathcal D_{\mathrm{Rx}}}
\!\left[\mathcal L_{\mathrm{RA}}\!\left(
\mathcal R_{\boldsymbol\vartheta}(\mathbf Y_p,\mathbf X_p;\Ksf),
\mathcal T\right)\right].
\label{eq:rx_adaptation}
\end{equation}

The pooled distribution of cached training data
$\mathcal D_{\mathrm{Rx}}$ is specified in
Section~\ref{sec:exp_setup}. Adaptation updates only
$\boldsymbol\vartheta$, detaching candidate choices and
assignments while retaining gradients through continuous
correction and physical refinement.

\setlength{\textfloatsep}{6pt plus 2pt minus 2pt}
\setlength{\dbltextfloatsep}{6pt plus 2pt minus 2pt}
\begin{table}[!t]
    \centering
    \caption{Main system and scene parameters.}
    \label{tab:params}
    \footnotesize
    \setlength{\tabcolsep}{4pt}
    \begin{tabular}{ll}
    \toprule
    Parameter & Value \\
    \midrule
    Carrier / sampling rate & 2.45 GHz / 150 MHz \\
    LFM bandwidth & 150 MHz \\
    Tx / Rx elements; spacing & $8/8$; $\lambda/2$ \\
    Frame pulses / pulse samples & $\Psf=16$, $\Lsf=256$ \\
    Spatial modes / observed pulses & $\Msf=4$, $\Psf_{\mathrm{obs}}=1$ \\
    Training target count; velocity & $\Ksf=3$; static \\
    Range / angle support & $[15,65]$ m / $[-60^\circ,60^\circ]$ \\
    Support of reflection amplitudes & $[0.5,1.0]$ \\
    Angle / delay points before masking & $64/60$ \\
    Valid angle / delay points & $55/50$ \\
    Image resolution & $256\times256$ \\
    Training SNR support & $[8,20]$ dB \\
    Reference sensing power & $P_{\mathrm{ref}}\approx1.9817$ \\
    \bottomrule
    \end{tabular}
\end{table}

\begin{table*}[!tb]
\centering
\caption{Waveform and UE configurations. The main waveform
comparisons use the baseline sensing DSP with each configuration's
complete realized waveform. Citations identify the underlying
principles of superposition or frequency separation.}
\label{tab:baselines}
\footnotesize
\setlength{\tabcolsep}{4pt}
\renewcommand{\arraystretch}{1.05}
\begin{tabular}{p{0.22\textwidth}p{0.40\textwidth}p{0.31\textwidth}}
\toprule
\textbf{Configuration} & \textbf{Waveform construction} & \textbf{UE configuration}\\
\midrule
Proposed & Temporal/spectral updates of a shared state
& Shared UE using reference and CSI\\
TD superposition~\cite{tagliaferri2023integrated} & Fixed source combination, $\rho=0.5$ & Same UE processing\\
Learned spectral linear & Complex source mixing in eight bands; 512 real parameters
& Same UE processing\\
Temporal nonlinear & Adaptive multiscale temporal features; no spectral path & Same UE processing\\
Fixed FDM with adapted UE~\cite{mao2022waveform} & Fixed complementary subbands and source filters
& Matched subband UE, adapted for a fixed Tx\\
Optimized FDM~\cite{mao2022waveform} & Four communication modes; sensing powers and phases learned on fixed subcarriers & Matched subband UE, jointly trained\\
\bottomrule
\end{tabular}
\end{table*}

\section{Experimental Results}
\label{sec:experiments}

We first compare waveform constructions and their sensitivity to noise.
We then hold the transmitted waveforms fixed to evaluate sequential
sensing recovery and the additional contribution of learned local
correction.

\subsection{Experimental Setup}
\label{sec:exp_setup}

\subsubsection{System, data, and metrics}
Channels follow $\mathbf h_{\mathrm c}\sim
\mathcal{CN}(\mathbf0,\mathbf I_{\Nsf_{\mathrm t}})$; three static targets
are sampled independently with $r_k\sim\mathcal U[15,65]$~m,
$\theta_k\sim\mathcal U[-60^\circ,60^\circ]$ and no minimum separation
(Table~\ref{tab:params}). The prescribed sensing covariance is constructed
over $[-45^\circ,45^\circ]$, whereas experimental target angles are sampled
over the wider interval $[-60^\circ,60^\circ]$. Real echo coefficients $\alpha_k\sim\mathcal U[0.5,1]$ remain constant within each frame, with no additional reflection phase or range attenuation. Each image uses 16 pulses, with each sensing decision
using one echo with 256 samples and its realized transmit pulse.
The communication receive SNR $\Gamma_{\mathrm c}$ and reference sensing
SNR $\Gamma_{\mathrm s}$ are expressed in dB, with nominal values of 10~dB.
Their noise variances are
\[
\sigma_{\mathrm c}^2=P_{\mathrm c}10^{-\Gamma_{\mathrm c}/10},\qquad
\sigma_{\mathrm s}^2=P_{\mathrm{ref}}10^{-\Gamma_{\mathrm s}/10},
\]
where $P_{\mathrm c}=(\Psf\Lsf)^{-1}\sum_p\|\mathbf h_{\mathrm c}^{T}\mathbf X_p\|_2^2$
is the noiseless power of the received frame. The sensing reference power
$P_{\mathrm{ref}}$ is fixed across methods and task weights. Thus,
communication noise follows each frame's received power, whereas sensing
noise at a given reference SNR is shared across waveforms.

The grid before masking contains 64 points for the sine of the angle with nominal
spacing $\Delta u=2/64$ from $-1$ to $1-2/64$
(first point clipped to $-0.999$), and 60 delay indices at unit spacing $10$--$69$. Physical masking retains 55 angle points
in $[-0.84375,0.84375]$ and 50 delay indices $16$--$65$.
Sequential recovery uses this retained grid.

We report peak signal-to-noise ratio (PSNR) and SSIM for image
reconstruction, and mean absolute error (MAE) in range and angle
for localization. Target estimates are paired with ground truth
by the one-to-one assignment in Section~\ref{sec:training}.
The normalized localization error is
\[
D_{\mathrm{RA}}=\tfrac12\left(
\frac{\mathrm{MAE}_{r}}{r_{\mathrm{ref}}}+
\frac{\mathrm{MAE}_{\theta}}{\theta_{\mathrm{ref}}}\right).
\]

Fixed scales calibrated on training data $r_{\mathrm{ref}}\approx0.3931$~m,
$\theta_{\mathrm{ref}}\approx1.326^\circ$, $n_{\mathrm c}\approx0.01886$,
and $n_{\mathrm s}\approx1.2992$ are shared across methods.
Relative reductions are computed from errors averaged equally over all
eleven weights; the same calculation is repeated within each paired bootstrap resample using images as clusters.

\subsubsection{Comparison design}
The proposed, linear, and temporal nonlinear transmitters share encoder
initialization and fixed UE processing, but initialize fusion operators
separately (Table~\ref{tab:baselines}). For fixed FDM, transmitter training computes the communication loss
through the same fixed UE receiver. The trained transmitter is then
held fixed while a UE matched to the subband is adapted; optimized FDM with four modes jointly trains its encoder, sensing resources, and matched UE. Both FDM receivers extract the
communication subband and correct channel gain before image recovery.
The proposed model and the linear controls share the source normalizations in
Sections~\ref{ssec:tx} and~\ref{subsec:image_representation}.
For the expanded sources defined in Section~\ref{subsec:fusion_motivation},
TD superposition~\cite{tagliaferri2023integrated} applies
\[
\mathbf S_{\mathrm{TD}}=\sqrt{1-\rho}\,\mathbf s+\sqrt{\rho}\,\mathbf z,
\qquad\rho=0.5,
\]
without further branch normalization.
With hats denoting $\mathcal F_{\Lsf}$ transforms, the spectral linear control applies learned source transforms:
\[
\widehat{\mathbf S}_{\mathrm{SL}}(:,p,f)=
\mathbf A_{\mathrm s,b(f)}\widehat{\mathbf s}(:,p,f)
+\mathbf A_{\mathrm c,b(f)}\widehat{\mathbf z}(:,p,f).
\]

For frequency $f$ in cycles/sample, $b(f)=1+\min(7,\lfloor16|f|\rfloor)$
defines eight bands of equal width in absolute frequency, each with complex
$\Msf\times \Msf$ matrices shared across pulses and bins. After inverse FFT
for spectral fusion, both controls use \eqref{eq:array_mapping}--\eqref{eq:array_waveform}
for spatial mapping and normalization of frame energy.
For FDM, frequency separation~\cite{mao2022waveform} is implemented here
with the centered half of the DFT bins assigned to communication and half
the energy of the array frame to each source, fixed across task weights.
Optimized FDM with four modes
filters the same image code's four cyclic modes into this subband and
maps them through $\mathbf B_{\mathrm{eff}}$ to one direction for maximum-ratio transmission (MRT) and three directions in the null space of the UE channel. Its three sensing codes use the leading covariance
eigenvectors, with learned subcarrier powers and phases in the
complementary subband. The separately trained temporal nonlinear control
retains temporal branches, adaptive joint weighting and physical
conditioning to assess the spectral extension. Fusion models were not
capacity matched.

\renewcommand{\topfraction}{0.99}
\renewcommand{\textfraction}{0.01}
\setlength{\floatsep}{2pt plus 1pt minus 1pt}
\setlength{\textfloatsep}{4pt plus 1pt minus 1pt}
\setlength{\dblfloatsep}{1pt plus 1pt minus 0pt}
\setlength{\dbltextfloatsep}{4pt plus 1pt minus 1pt}
\setlength{\abovecaptionskip}{3pt}
\clubpenalty=10000
\widowpenalty=10000

\begin{figure*}[!tb]
    \centering
    \includegraphics[width=0.85\textwidth]
    {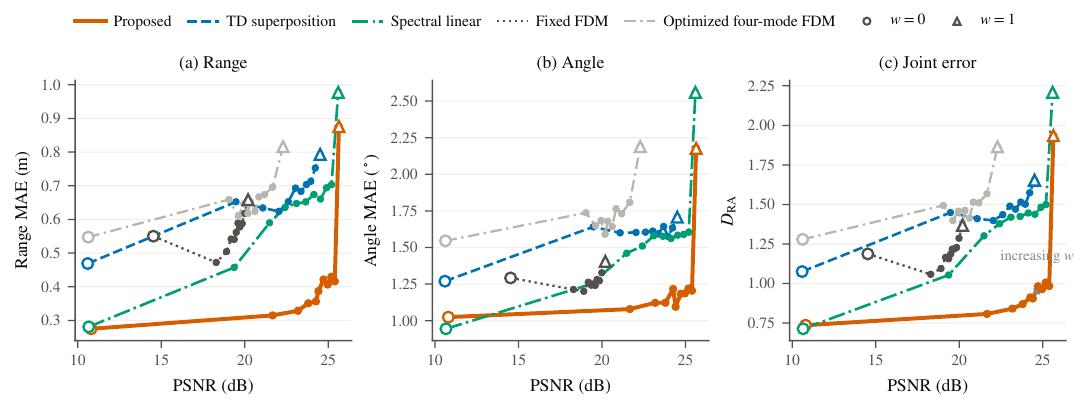}
    \caption{Tradeoffs on the development set between PSNR and (a) range MAE,
(b) angle MAE, and (c) normalized localization error.
Curves span eleven task weights; open circles and triangles mark
$w=0$ and $w=1$, respectively. All use the baseline DSP receiver.
The three methods other than FDM share the UE receiver; TD uses $\rho=0.5$,
and FDM uses UEs matched to their respective configurations.}
    \label{fig:sensing_comm_tradeoff}
\end{figure*}

\begin{figure}[!t]
    \centering
    \includegraphics[width=\columnwidth]
    {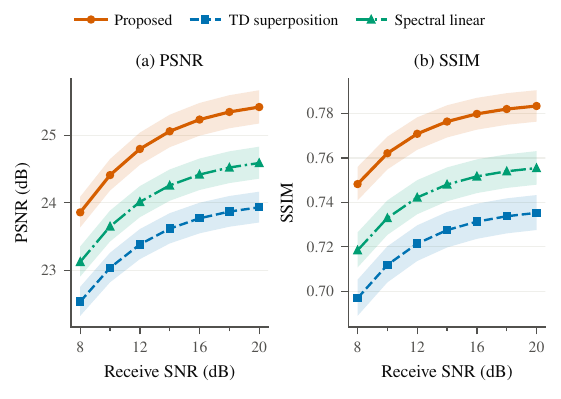}
    \caption{Image reconstruction on the development set: (a) PSNR and (b) SSIM versus receive SNR
at $w=0.5$, with the common UE receiver. Shading indicates
95\% bootstrap intervals using images as clusters, conditional on the trained models.}
    \label{fig:comm_vs_snr}
\end{figure}

\begin{figure}[!htb]
\centering
\includegraphics[width=0.95\columnwidth]{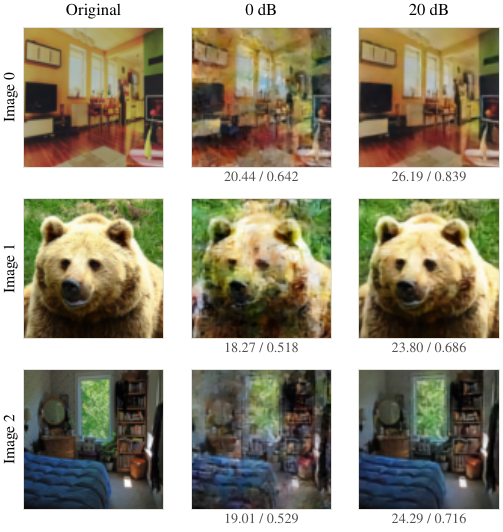}
\caption{Original images and Proposed reconstructions for fixed development
images 0--2 at $w=0.5$. Reconstruction columns use 0 and 20~dB
receive SNR; labels give PSNR (dB)/SSIM.}
\label{fig:recon_visual}
\end{figure}

\subsubsection{Training and evaluation}
\label{subsec:training_evaluation}
The shared UE and encoder initialization came from the final exponentially averaged parameters of a reference transceiver with four modes, conditioned on LFM and trained for 60 epochs. Its complex temporal operator fused expanded image and LFM features using CSI and covariance, followed by spatial mapping and normalization of frame energy.
Reference pretraining combined losses for image reconstruction and LFM ambiguity; an auxiliary heatmap/coordinate sensing loss was introduced
after 20 epochs. The encoder, fusion operator and UE were trained
throughout, with the auxiliary sensing receiver also trained from
epoch 21. All
constants for loss normalization and the reference sensing power were
calibrated on training data and then fixed across waveform methods and
task weights.

Starting from the common encoder initialization,
waveform training used Adam, batches of 32 frames,
and 60 epochs on a COCO stream of 10\,000 images~\cite{lin2014microsoft}.
Epochs 1--30 used $\Qsf=1$ and peak learning rate $10^{-4}$;
epochs 31--60 used $\Qsf=4$ and $2.5\times10^{-5}$.
Each training phase used three warmup epochs followed by cosine
decay, with respective floors $2.5\times10^{-5}$ and $10^{-6}$.
Temperature was fixed at $T=0.025$. Residual scales in the joint blocks were
initialized to $0.28$, and the four bandwise spectral scales to
$(\kappa_1,\kappa_2,\kappa_3,\kappa_4)=(0.10,0.12,0.14,0.16)$.

\begin{figure*}[!tb]
\centering
\setlength{\abovecaptionskip}{2pt}
\includegraphics[
    width=0.85\textwidth,
    trim=0bp 7.5bp 0bp 4.5bp,
    clip
]{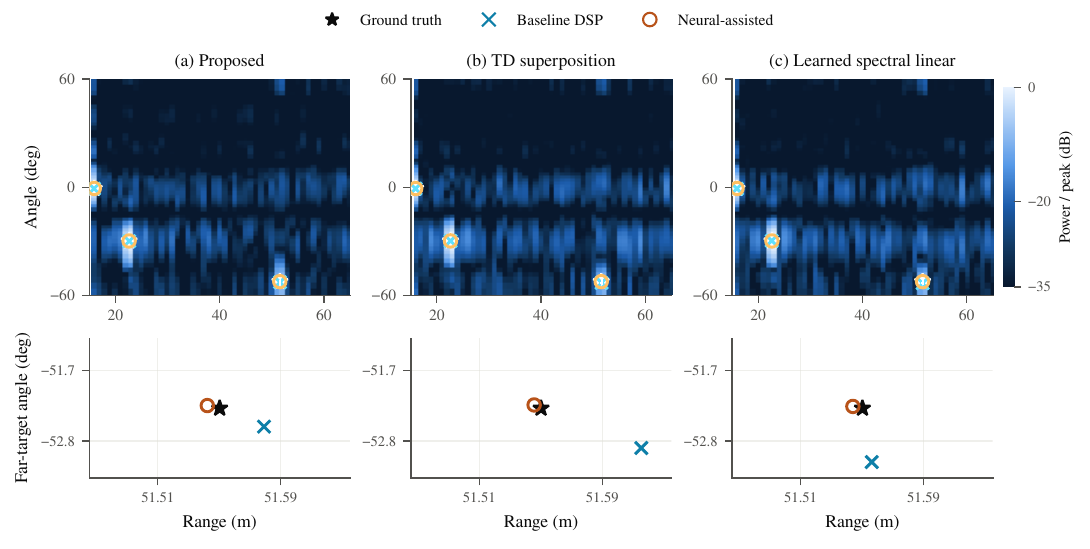}
\caption{Fixed development example: scene/image~0, pulse~0, $w=0.5$,
10~dB reference sensing SNR, receiver seed~42.
Columns compare Proposed, TD superposition, and learned spectral linear.
Top: coherent scores normalized by their peak, with ground truth and estimates
from baseline DSP and recovery with neural assistance. Bottom: a common zoom around the farthest target.}
\label{fig:target_comparison}
\end{figure*}

With the trained transmitters fixed, one local network
per receiver seed was adapted for eight epochs on
COCO train2017 caches balanced across eleven weights and the four
configurations other than FDM in Table~\ref{tab:baselines}.
The same network served all these configurations, including
Fig.~\ref{fig:target_comparison}.

Evaluation used $w\in\{0,0.1,\ldots,1\}$, 512 COCO val2017 development
images and 512 test2017 images unused in training or method selection.
Models and evaluation rules were fixed before testing, with scenes,
channels and noise paired across methods. Each image supplies 16 separate
decisions based on individual pulses; SNR sweeps and examples use $w=0.5$.
Waveform models use one training seed; results with neural assistance average three receiver training seeds unless specified. Intervals use paired bootstrap resampling with images as clusters, conditional on trained models.

\subsection{Gains from Nonlinear Waveform Fusion}

\begin{figure}[!tb]
    \centering
    \includegraphics[width=0.90\columnwidth,trim=0bp 2bp 0bp 5bp,clip]
    {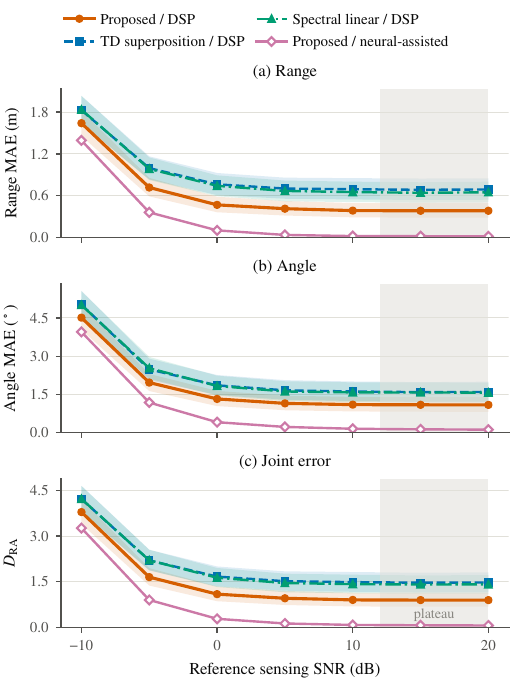}
    \caption{Localization on the development set: (a) range MAE, (b) angle MAE, and (c) normalized localization error versus reference sensing SNR at $w=0.5$ and $\Ksf=3$.
DSP denotes baseline coherent peak search; neural assistance denotes
sequential recovery with local correction. Colored bands are conditional
95\% bootstrap intervals; gray shading marks the region of high SNR.}
    \label{fig:sens_k3}
\end{figure}

\begin{figure}[!tb]
\centering
\includegraphics[width=0.90\columnwidth]{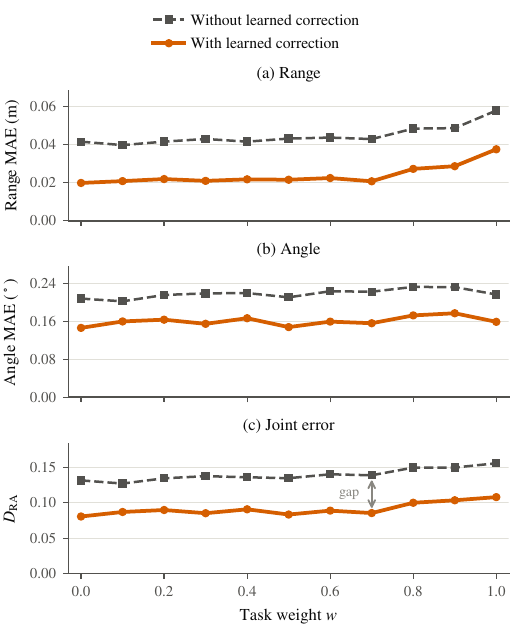}
\caption{Ablation of local correction with the proposed waveforms fixed:
(a) range MAE, (b) angle MAE, and (c) normalized localization error.
Both sequential variants share quadratic initialization and one physical
refinement per target addition. Points average 512 development images
at 10~dB reference sensing SNR; the corrected curve also averages
three seeds for receiver training.}
\label{fig:receiver_contribution}
\end{figure}

\begin{figure*}[!t]
\centering
\setlength{\abovecaptionskip}{2pt}
\includegraphics[
    width=0.85\textwidth,
    trim=0bp 7.5bp 0bp 8.5bp,
    clip
]{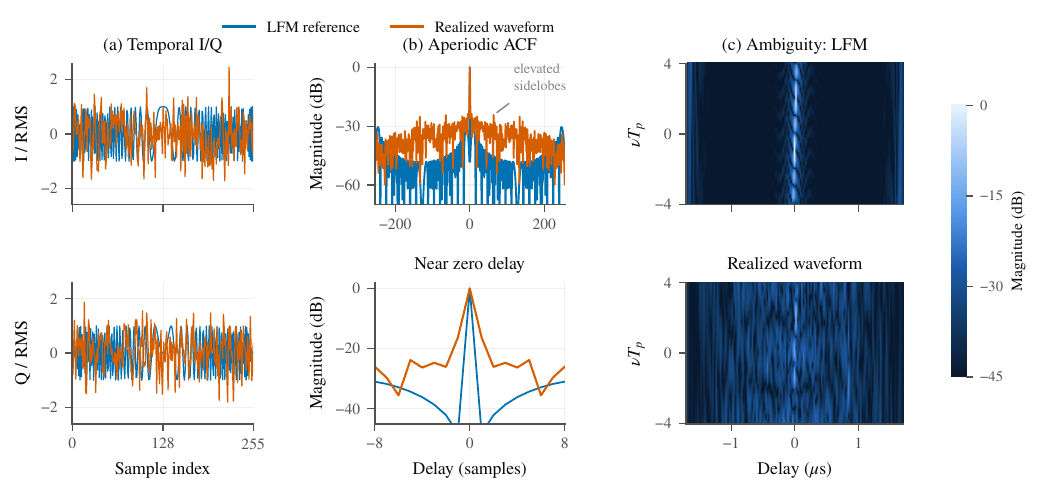}
\caption{LFM reference and realized proposed waveform (image~0, pulse~0):
(a) I/Q from antenna 0 normalized by each waveform's RMS;
(b) aperiodic autocorrelation and a zoom around zero delay;
(c) ambiguity magnitude. Correlations are summed coherently across
antennas and normalized by total pulse energy.
Ambiguity modulates the undelayed signal by $e^{-j2\pi\nu\ell/f_{\mathrm s}}$;
$\nu$ is Doppler frequency and $T_p=\Lsf/f_{\mathrm s}$ is pulse duration.
LFM has no spatial mapping. Localization uses the model with circular blocks in Section~\ref{ssec:sens}.}
\label{fig:waveform_analysis}
\end{figure*}

\label{sec:waveform_performance}
\label{subsec:tradeoff}

Figure~\ref{fig:sensing_comm_tradeoff} compares operating points for five waveforms on the development set, connecting models trained separately at eleven task weights.
The proposed construction achieves higher PSNR and lower
normalized localization error than each linear control
at most matched weights. Relative to fixed FDM and optimized FDM with four modes, evaluated with their respective
UE receivers matched to their subbands, both metrics improve at
nine and ten weights, respectively. At $w=1$, the localization
term vanishes from~\eqref{eq:task_formulation} and localization error rises sharply,
highlighting the role of sensing supervision in waveform learning.

Independent testing on 512 previously unused images confirms the joint
improvement under the common UE and baseline sensing DSP. With means
weighted equally over all eleven task weights, the proposed waveform reduces
$D_{\mathrm{RA}}$ by 29.0\% versus TD superposition (95\% confidence interval:
[20.9\%,36.6\%]) and 27.0\% versus learned spectral linear fusion
(95\% confidence interval: [19.2\%,34.4\%]). Mean PSNR gains are 1.40 and
0.82~dB, respectively. Both metrics improve at ten of eleven matched weights
against each control.

To assess the spectral extension, we compare the full model with
independently trained temporal nonlinear fusion under the same UE receiver
and baseline sensing DSP. On the development set, mean PSNR remains nearly
unchanged, at 23.093 and 23.094~dB for the temporal and full models,
respectively. Mean $D_{\mathrm{RA}}$ decreases from 1.0015 to 0.9950,
a reduction of approximately 0.65\%, and the normalized task objective
improves at five of eleven weights. The spectral extension thus provides
a modest increment varying with task weight in \mbox{this comparison}.

\subsection{Noise Sensitivity of the Learned Waveforms}
\label{sec:exp_comm}

With the $w=0.5$ models held fixed, reconstruction quality increases with
receive SNR and the proposed waveform maintains higher mean PSNR/SSIM than both
linear controls throughout the sweep (Fig.~\ref{fig:comm_vs_snr}).
Its PSNR gain over spectral linear is 0.829~dB at 20~dB.
Figure~\ref{fig:recon_visual} compares original images with reconstructions
at 0 and 20~dB. At 0~dB, scene layout remains recognizable but texture
and edges are less distinct; at 20~dB, sharper boundaries and detail
accompany higher PSNR and SSIM.

Under the baseline DSP, the proposed waveform achieves lower range,
angle, and normalized localization errors than both linear controls
throughout the tested reference sensing SNR range
(Fig.~\ref{fig:sens_k3}). All three DSP curves approach a plateau at high
SNR. For the same proposed waveform, sequential recovery with neural assistance
lowers all three errors and markedly reduces this plateau.
Section~\ref{sec:exp_sens} separates the gains from sequential recovery
and learned local correction.

\subsection{Contribution of Sequential Recovery and Local Learning}
\label{sec:exp_sens}

Figure~\ref{fig:receiver_contribution} isolates local correction by holding
the proposed waveforms, sequential procedure, quadratic initialization and
physical refinement fixed. On the development set, enabling
correction reduces range, angle and normalized localization errors at
every tested weight.

On the independent test set, mean $D_{\mathrm{RA}}$ over eleven
equally weighted task weights is 1.0899 for the baseline DSP receiver,
0.1185 for sequential recovery without correction, and 0.1041 with
correction. Amplitude refitting, physical refinement and residual
updates account for most of the reduction. Local correction of
candidate initialization further reduces error by 12.2\%
(95\% CI: [5.8\%,19.0\%]), improving every weight with a smaller
aggregate gain than on the development set. Communication outputs
are unchanged.

\par

In the common zoom around the farthest target in Fig.~\ref{fig:target_comparison},
estimates with neural assistance lie closer to ground truth for all
three waveforms, showing coordinate improvements at a scale finer than the grid.

\par

\subsection{Waveform Structure and Correlation Diagnostics}
\label{sec:exp_waveform_analysis}

Figure~\ref{fig:waveform_analysis} compares the LFM reference
with a realized transmit pulse. The latter retains a distinct autocorrelation peak at zero delay, but exhibits higher sidelobes
over much of the delay range and a different ambiguity pattern.
This behavior is consistent with waveform optimization
under~\eqref{eq:task_formulation}, which uses image reconstruction
and localization losses without an explicit penalty on waveform similarity or sidelobes. The coherent dictionary
in~\eqref{eq:echo_atom} incorporates the realized pulse and its structure determined by image content.

\section{Conclusion}
\label{sec:conclusion}

This paper developed nonlinear waveform fusion across scales for ISAC by learning interactions between image features and a deterministic LFM reference. The UE reconstructs images using the reference and CSI; the BS localizes multiple targets in range and angle using the realized pulse.

On independent test images, the proposed waveforms achieve higher PSNR and
lower normalized localization error at ten of eleven task weights against each
linear control under the shared fixed UE receiver and baseline DSP
receiver. Relative to TD superposition and learned spectral linear fusion,
mean localization error decreases by 29.0\% and 27.0\%, with corresponding
PSNR gains of 1.40 and 0.82~dB. With the proposed waveforms fixed,
sequential physical recovery substantially reduces localization error,
and learned local correction yields a further 12.2\% reduction.
These results show how synthesis adapted to image content and
sequential recovery using the transmitted waveform complement each other
for image transmission and localization of multiple targets.

\IEEEtriggeratref{0}
\bibliographystyle{IEEEtran}
	\bibliography{reference}
\end{document}

%% file: macros.tex
\long\def\comment#1{}

\newcommand{\Csf}{{\mathsf C}}
\newcommand{\Dsf}{{\mathsf D}}

\newcommand{\Ksf}{{\mathsf K}}
\newcommand{\Lsf}{{\mathsf L}}
\newcommand{\Msf}{{\mathsf M}}
\newcommand{\Nsf}{{\mathsf N}}

\newcommand{\Psf}{{\mathsf P}}
\newcommand{\Qsf}{{\mathsf Q}}

\renewcommand{\arg}{{\hbox{arg}}}

\renewcommand{\Re}{{\rm Re}}
\renewcommand{\Im}{{\rm Im}}

\providecommand{\definitionname}{Definition}